%% file: TL-Align.tex
\documentclass[10pt,twocolumn,letterpaper]{article}

\makeatletter
\@namedef{ver@everyshi.sty}{}
\makeatother

\usepackage[pagenumbers]{cvpr} 

\usepackage{graphicx}
\usepackage{amsmath}
\usepackage{amssymb}
\usepackage{booktabs}
\usepackage{tabu}
\usepackage{adjustbox}
\usepackage{multirow}

\usepackage{url}

\usepackage{todonotes}
\usepackage{float}

\usepackage{mathtools}

\usepackage{amsthm}

\usepackage[pagebackref,breaklinks,colorlinks]{hyperref}

\usepackage[capitalize]{cleveref}
\crefname{section}{Sec.}{Secs.}
\Crefname{section}{Section}{Sections}
\Crefname{table}{Table}{Tables}
\Crefname{equation}{}{}

\def\cvprPaperID{****} 
\def\confName{CVPR}
\def\confYear{2023}

\begin{document}

\title{Token-Label Alignment for Vision Transformers}

\newcommand*\samethanks[1][\value{footnote}]{\footnotemark[#1]}
\author{Han Xiao$^{1,2,}$\footnotetext{sadfsadfa}\thanks{Equal contribution.}\quad Wenzhao Zheng$^{1,2,}$\samethanks\quad Zheng Zhu$^3$\quad Jie Zhou$^{1,2}$\quad Jiwen Lu$^{1,2,}$\thanks{Corresponding author.} \\
$^1$Beijing National Research Center for Information Science and Technology, China \\
$^2$Department of Automation, Tsinghua University, China \quad\quad $^3$PhiGent Robotics \\
\texttt{\{h-xiao20,zhengwz18\}@mails.tsinghua.edu.cn; zhengzhu@ieee.org;} \\
\texttt{\{jzhou,lujiwen\}@tsinghua.edu.cn}
}

\maketitle

\input{chapters/0_abstract.tex}
\input{chapters/1_introduction.tex}

\input{chapters/2_related_work.tex}

\input{chapters/3_proposed_approach.tex}

\input{chapters/4_experiment.tex}

\input{chapters/5_conclusion.tex}

\appendix

\input{chapters/a_training.tex}
\input{chapters/a_exp_details.tex}

\input{chapters/a_vis.tex}
\input{chapters/a_generalize.tex}

{\small

\input{reference.bbl}
\bibliographystyle{ieee_fullname}
}

\end{document}

%% file: chapters/0_abstract.tex
\begin{abstract}
Data mixing strategies (e.g., CutMix) have shown the ability to greatly improve the performance of convolutional neural networks (CNNs).
They mix two images as inputs for training and assign them with a mixed label with the same ratio.
While they are shown effective for vision transformers (ViTs), we identify a token fluctuation phenomenon that has suppressed the potential of data mixing strategies.
We empirically observe that the contributions of input tokens fluctuate as forward propagating, which might induce a different mixing ratio in the output tokens.
The training target computed by the original data mixing strategy can thus be inaccurate, resulting in less effective training.
To address this, we propose a token-label alignment (TL-Align) method to trace the correspondence between transformed tokens and the original tokens to maintain a label for each token.
We reuse the computed attention at each layer for efficient token-label alignment, introducing only negligible additional training costs. 
Extensive experiments demonstrate that our method improves the performance of ViTs on image classification, semantic segmentation, objective detection, and transfer learning tasks. 
Code is available at: \url{https://github.com/Euphoria16/TL-Align}.
\end{abstract}

%% file: chapters/1_introduction.tex
\vspace{-5mm}
\section{Introduction} 
The recent developments of vision transformers (ViTs) have revolutionized the computer vision field and set new state-of-the-arts in a variety of tasks, such as image classification~\cite{dosovitskiy2020image,touvron2021training,liu2021swin, chu2021twins}, object detection~\cite{carion2020end,zhu2020deformable,dai2021dynamic,dai2021up}, and semantic segmentation~\cite{li2017not,strudel2021segmenter,zheng2021rethinking,cheng2021per}.
The successful structure of alternative spatial mixing and channel mixing in ViTs also motivates the arising of high-performance MLP-like deep architectures~\cite{tolstikhin2021mlp, touvron2021resmlp, tang2021image,wei2022activemlp} and promotes the evolution of better CNNs~\cite{ding2022scaling,liu2022convnet,guo2022visual}.
In addition to architecture designs, an improved training strategy can also greatly boost the performance of a trained deep model~\cite{jiang2021all,touvron2022deit,chen2022principle,chen2021transmix}.

The training of modern deep architecture almost all adopts data mixing strategies for data augmentation~\cite{walawalkar2020attentive,uddin2020saliencymix,kim2020puzzle,verma2019manifold,yun2019cutmix,zhang2018mixup}, which have been proven to consistently improve the generalization performance.
They randomly mix two images as well as their labels with the same mixing ratio to produce mixed data.
As the most commonly used data mixing strategy, CutMix~\cite{yun2019cutmix} performs a copy-and-paste operation on the spatial domain to produce spatially mixed images.
While data mixing strategies have been widely studied for CNNs~\cite{walawalkar2020attentive,uddin2020saliencymix,kim2020puzzle}, few works have explored their compatibilities with ViTs~\cite{chen2021transmix}.
We find that self-attention in ViTs causes a fluctuation of the original spatial structure.
Unlike the translation equivalence that ensures a global label consistency for CNNs, self-attention in ViTs undermines this global consistency and causes a misalignment between the token and label.
This misalignment induces a different mixing ratio in the output tokens.
The training targets computed by the original data mixing strategies can then be inaccurate, resulting in less effective training.

\begin{figure*}[t]
	\centering
	\includegraphics[width=\textwidth]{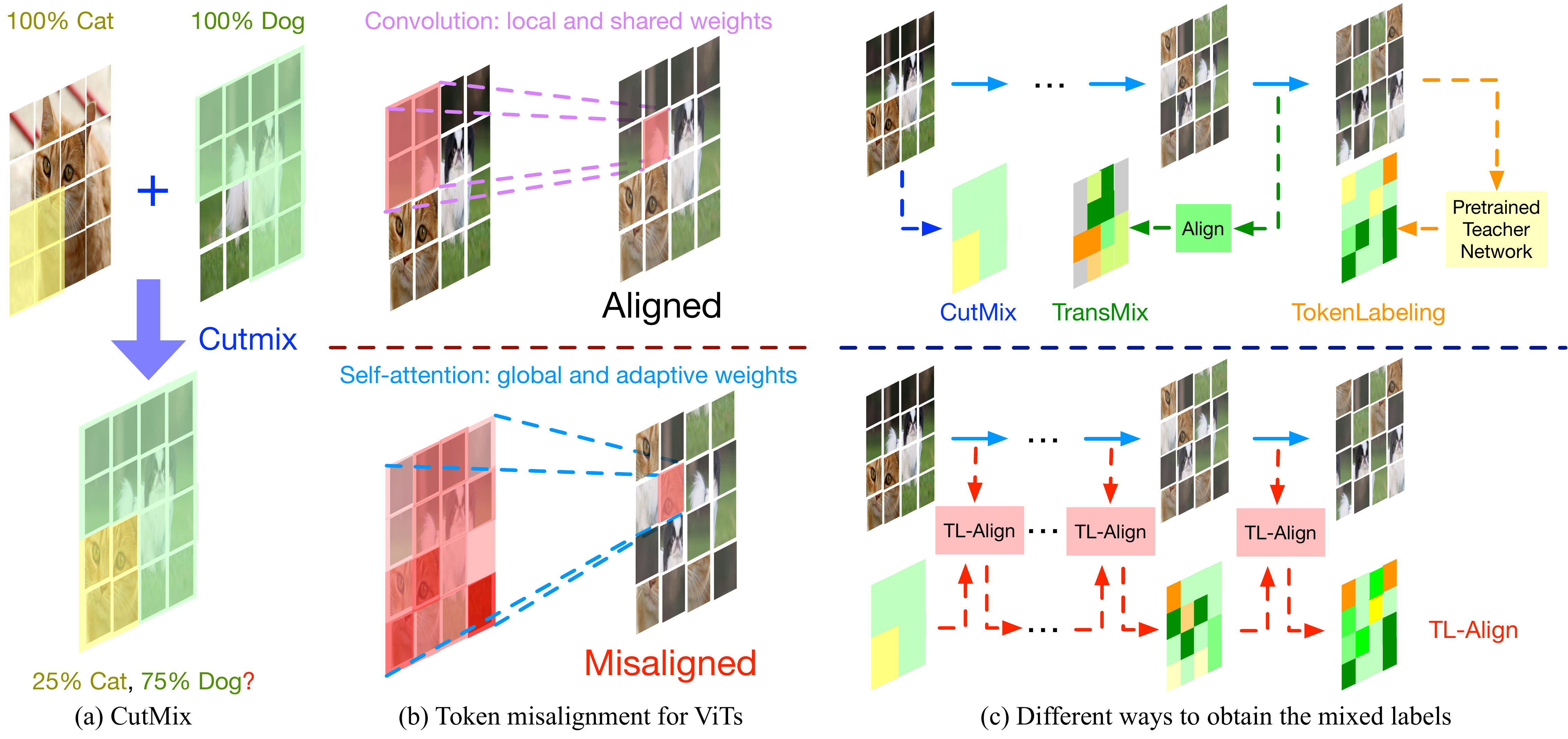}
\vspace{-7mm}
  \caption{
An overview of the proposed TL-Align. 
(a) CutMix-like methods~\cite{yun2019cutmix} are widely used in model training, which spatially mix the tokens and their labels in the input space. 
(b) They are originally designed for CNNs and assume the processed tokens are spatially aligned with the input tokens. 
We show that it does not hold true for ViTs due to the global receptive field and the adaptive weights.
(c) Compared with existing methods, our method can effectively and efficiently align the tokens and labels without requiring a pretrained teacher network.
	  } 
  \label{fig:overview}
 \vspace{-6mm}
 \end{figure*}

To address this, we propose a token-label alignment (TL-Align) method for ViTs to obtain a more accurate target for training.
We present an overview of our method in \Cref{fig:overview}.
We first assign a label to each input token in the mixed image according to the source of the token.
We then trace the correspondence between the input tokens and the transformed tokens and align the labels accordingly.
We assume that only the spatial self-attention and residual connection operation alter the presence of input tokens since channel MLP and layer normalization process each token independently.
We reuse the computed attentions to linearly mix the labels of input tokens to obtain those of transformed tokens.
The token-label alignment is performed iteratively to obtain a label for each output token.
For class-token-based classification (e.g., ViT~\cite{dosovitskiy2020image} and DeiT~\cite{touvron2021training}), we directly use the aligned label for the output class token as the training target.
For global-pooling-based classification (e.g., Swin~\cite{liu2021swin}), we similarly average the labels of output tokens as the training target.
The proposed TL-Align is only used for training to improve performance and introduces no additional workload for inference.
We apply the proposed TL-Align to various ViT variants with CutMix including plain ViTs (DeiT~\cite{touvron2021training}) and hierarchical ViTs (Swin~\cite{liu2021swin}).
We observe a consistent performance boost across different models on ImageNet-1K~\cite{deng2009imagenet}.
Specifically, our TL-Align improves DeiT-S by 0.8\% using the same training recipe.
We evaluate the ImageNet-pretrained models on various downstream tasks including semantic segmentation, objection detection, and transfer learning.
Experimental results also verify the robustness and generalization ability of our method.

%% file: chapters/2_related_work.tex
\section{Related Work}

\textbf{Vision Transformer.}
Transformers have been widely used in natural language processing and achieved great success on many language tasks. 
Recently, Vision Transformers (ViTs) have aroused extensive interest in computer vision due to their competitive performance compared with CNNs~\cite{dosovitskiy2020image,touvron2021training,liu2021swin, chu2021twins}. 
Dosovitskiy~et al.~\cite{dosovitskiy2020image} firstly introduced transformers into the image classification task.
They split the input image into non-overlapped patches and then feed them into the transformer encoders. 
Liu~et al.~\cite{liu2021swin} proposed a shifted windowing scheme to produce hierarchical feature maps suitable for dense prediction tasks. 
The great potential of vision transformer has motivated its adaptation to many challenging tasks including object detection~\cite{dai2021dynamic,zhu2020deformable,carion2020end}, segmentation~\cite{cheng2021per,strudel2021segmenter}, image enhancement~\cite{chen2021pre,li2021efficient} and video understanding~\cite{liu2021video,arnab2021vivit}. 

Recently, some efforts have been devoted to producing better training targets to improve the performance of vision transformers~\cite{jiang2021all,touvron2022deit}. 
For example, DeiT~\cite{touvron2021training} introduces a knowledge distillation procedure to reduce the training cost of ViTs and achieves a better accuracy/speed trade-off. 
TokenLabeling~\cite{jiang2021all} employs a pretrained teacher annotator to predict a label for each token for dense knowledge distillation.
Differently, we do not require a pretrained network to obtain the training targets.
Our TL-Align maintains an aligned label for each token layer by layer and can be trained efficiently in an end-to-end manner.

\textbf{Data Mixing Strategy.}
As an important type of data augmentation, data mixing strategies have demonstrated a consistent improvement in the generalization performance of CNNs. 
Zhang~et al.~\cite{zhang2018mixup} first proposed to combine a training pair to create augmented samples for model regularization. 
They perform linear interpolations on both the input images and associated targets. 
Following MixUp, CutMix~\cite{yun2019cutmix} also utilizes the mixture of two input images but adopts a region copy-and-paste operation. 
Later methods including Puzzle Mix~\cite{kim2020puzzle}, SaliencyMix~\cite{uddin2020saliencymix} and Attentive CutMix~\cite{walawalkar2020attentive} leverage the salient regions for informative mixture generation. 
Recently, Yang~et al.~\cite{yang2022recursivemix} proposed a RecursiveMix strategy which employs the historical input-prediction-label triplets for scale-invariant feature learning. 
Despite the better performance, a drawback of these methods is the heavily increased training cost due to the saliency extraction or historical information exploitation.

Most existing data mixing methods are originally designed for CNNs, and their effectiveness on ViTs has not been well explored.
TransMix~\cite{chen2021transmix} uses the class attention map at the last layer to re-weight the mixing targets and assumes the output tokens to keep spatial correspondence with the input tokens.
However, we identify a token fluctuation phenomenon for ViTs which may cause a mismatch between tokens and labels, leading to inaccurate label assignments in both the original CutMix and TransMix.
To address this, we propose to align the label and token space by tracing their correspondence in a layerwise manner.

%% file: chapters/3_proposed_approach.tex
\section{Proposed Approach}

\subsection{Preliminaries}
The convolution neural network (CNN) has been the dominant architecture for computer vision in the deep learning era, greatly improving the performance of many tasks. 
Its monopoly has been challenged by the recent emergence of vision transformers (ViTs), which first ``patchify'' each image into tokens and process them with alternating self-attention (SA) and multi-layer perceptron (MLP). 

In addition to architecture design, training strategy also has a large effect on the model performance, especially the data augmentation strategy.
Data mixing~\cite{walawalkar2020attentive,uddin2020saliencymix,kim2020puzzle,verma2019manifold,yun2019cutmix,zhang2018mixup} is an important set of data augmentation for the training of both CNNs and ViTs, as it significantly improves the generalization ability of models. 
As the most commonly used data mixing strategy, CutMix~\cite{yun2019cutmix} aims to create virtual training samples from the given training samples $(\mathbf{X},y)$, where $\mathbf{X} \in \mathcal{R}^{H\times W \times C}$ denotes the input image and $y$ is the corresponding label. 
CutMix randomly selects a local region from one input $\mathbf{X}_1$ and uses it to replace the pixels in the same region of another input $\mathbf{X}_2$ to generate a new sample $\mathbf{\mathbf{\tilde X}}$. 
Similarly, the label $\tilde y$ of $\mathbf{\mathbf{\tilde x}}$ is also the combination of the original labels $y_1$ and $y_2$:
\begin{equation}
\begin{aligned}
    \mathbf{\mathbf{\tilde X}} &= \mathbf{M} \odot {\mathbf{X}_1} +(\mathbf{1}-\mathbf{M}) \odot {\mathbf{X}_2} \\
    \tilde y &= \lambda y_1 + (1-\lambda) y_2
\end{aligned}
\label{eq:cutmix}
\end{equation}
where $M\in \{0,1\}^{H\times W}$ is a binary mask indicating the image each pixel belongs to, $\mathbf{1}$ is an all-one matrix, and $\odot$ is the element-wise multiplication.
 $\lambda$ reflects the mixing ratio of two labels and is the proportion of pixels cropped from $\mathbf{X}_1$ in the mixed image $\mathbf{\mathbf{\tilde X}}$. For a cropped region $[r_x, r_x+r_w]\times [r_y, r_y+ r_h]$ from $\mathbf{X}_1$, we compute $\lambda = \frac{r_w r_h}{WH}$ to obtain the initial mixed target $\tilde y$.

\subsection{The Token Fluctuation Phenomenon} \label{sec:issue}
CutMix is originally designed for CNNs and assumes the feature extraction process does not alter the mixing ratio.
However, we discover that different from CNNs, self-attention in ViTs can lead to the fluctuation of some tokens.
The fluctuation further results in the mismatch between the token space and label space, which hinders the effective training of the network.

Formally, we use $\mathbf{z}_i$ to denote a token of the image $\mathbf{Z}$, i.e., $\mathbf{z}_i$ is the transposed $i$-th column vector of $\mathbf{Z}$.
We can then compute the $i$-th transformed token $\hat{\mathbf{z}}_i$ after the spatial operation as $\hat{\mathbf{z}}_i = \sum_{j=1}^{N} w^s_{i,j} \mathbf{z}_j$,
where $w^s_{i,j}$ is the $i,j$-th element of the computed spatial mixing matrix $\mathbf{w}^s(\mathbf{z})$.

With the assumption of the linear information integration, we define the contribution of an original token $\mathbf{z}_i$ to a mixed token $\hat{\mathbf{z}}_j$ as $c(\mathbf{z}_i, \hat{\mathbf{z}}_j) = \frac{|w^s_{i,j}|}{\sum_{k=1}^{N} |w^s_{k,j}|},$
where $|\cdot|$ denotes the absolute value.
We can then compute the presence of a token $\mathbf{z}_i$ in all the mixed image tokens as:
\begin{equation} \label{eq:contri_all}
p (\mathbf{z}_i) = \sum_{j=1}^{N} c(\mathbf{z}_i, \hat{\mathbf{z}}_j) = \sum_{j=1}^{N} \frac{|w^s_{i,j}|}{\sum_{k=1}^{N} |w^s_{k,j}|}.
\end{equation}

For non-strided depth-wise convolution, each token is multiplied by each element in the convolutional kernel due to the translation invariance. We can thus obtain:
\begin{equation} \label{eq:contri_conv}
\sum_{l=1}^{N} |w^s_{i,l}| = \sum_{j=1}^{N} |w^s_{k,j}| = \sum_{k=1,l=1}^{M} |K_{k,l}|, \ \ \ \ \  \forall i,j \in \mathbf{P}_{NE},
\end{equation}
where $\mathbf{P}_{NE}$ denotes the set of positions that are not at the edge of the image, $K_{k,l}$ denotes the value of the $k,l$-th position of the convolution kernel $\mathbf{K}$ and $M$ is the kernel size.
We can infer that $p (\mathbf{z}_i) = 1, \ \forall i \in \mathbf{P}_{NE}$, i.e., the effect of all the internal tokens does not change during the convolution process.
However, for self-attention in ViTs, \cref{eq:contri_conv} does not hold due to the non-existence of translation invariance.
The fluctuation of $p (\mathbf{z})$ is further amplified by the input dependency of the spatial mixing matrix $\mathbf{w}^s(\mathbf{z})$ induced by self-attention.
As an extreme case, we may obtain $p(\mathbf{z}) \sim 0$ for certain tokens.
The fluctuation of tokens will alter the proportion of mixing (i.e., $\lambda$) and the network might even completely ignore one of the mixed images.
The actual label of the processed tokens can then deviate from the mixed label computed by \cref{eq:cutmix}, resulting in less effective training.

\begin{figure*}[t]
	\centering
	\includegraphics[width=0.9\textwidth]{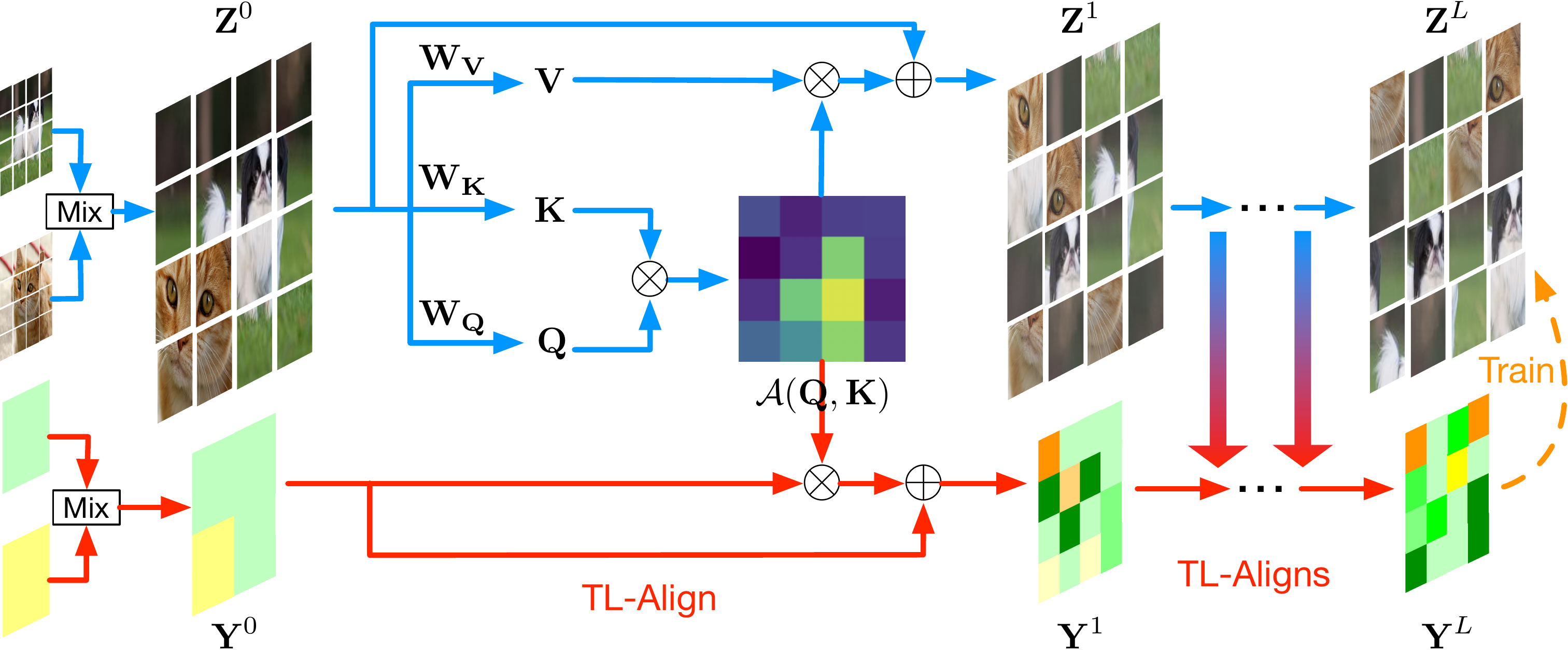}
  \vspace{-1mm}
  \caption{Illustration of the proposed TL-Align.
  We trace the correspondence between the input tokens and the transformed tokens and align the labels accordingly.
  We reuse the computed attentions to linearly mix the labels of input tokens to obtain those of transformed tokens.
The token-label alignment is performed iteratively to obtain a label for each output token.
	  } 
  \label{fig:framework}
  \vspace{-6mm}
 \end{figure*}

\subsection{Token-Label Alignment}
Each token in ViTs interacts with other tokens using the self-attention mechanism.
The input-dependent weights empower ViTs with more flexibility but also result in a mismatch between the processed token and the initial token.
To address this, we propose a token-label alignment (TL-Align) method to trace the correspondence between the input and transformed tokens to obtain the aligned labels for the resulting representations, as illustrated in \Cref{fig:framework}.

Specifically, ViTs first split the mixed input $\mathbf{\tilde X}$ after CutMix \cref{eq:cutmix} to a sequence of $N$ non-overlapped patches and then flatten them to obtain the original image tokens $\{ \mathbf{\tilde x}_1, \mathbf{\tilde x}_2, \cdots, \mathbf{\tilde x}_N  \}$.
We then project them into a proper dimension and add positional embeddings:
\begin{equation} \label{eq:toikenize}
\mathbf{Z}^{0} = [{\mathbf{\mathbf{\tilde z}}}_{cls};{\mathbf{\mathbf{\tilde x}}}_{1} \cdot \mathbf{E};{\mathbf{\mathbf{\tilde x}}}_{2} \cdot \mathbf{E}; \cdots ;{\mathbf{\mathbf{\tilde x}}}_{N} \cdot \mathbf{E}] + \mathbf{E}_{pos},
\end{equation}
where ${\mathbf{\tilde z}}_{cls} \in \mathcal{R}^{1\times d}$ denotes the class token, N is the number of tokens, $\mathbf{E}$ represents the patch projector, and $\mathbf{E}_{pos} \in \mathcal{R}^{(N+1)\times d}$ is the position embeddings.
Note that we adopt the process of the original transformer architecture~\cite{dosovitskiy2020image} as an example without loss of generality. 
Other models may omit the class token and use a relative positional embedding instead, which does not affect the utility of the proposed TL-Align method.

We first assign each token $\mathbf{z}_i \in \mathcal{R}^{1\times d}$ with a label embedding $\mathbf{y}_i \in \mathcal{R}^{1\times C}$:
\begin{equation}\label{eq:input_y}
\mathbf{Y}^{0} = [\tilde{\mathbf{y}}^{0}_{cls};\tilde{\mathbf{y}}^{0}_{1};\tilde{\mathbf{y}}^{0}_{2};...;\tilde{\mathbf{y}}^{0}_{N}],
\end{equation}
where the sum of elements in each $\mathbf{y}_i$ equals 1 (i.e., $\sum_{j=1}^{C} y_{i,j} = 1$) and $y_{i,j}$ indicates how much the $i$-th token belong to the $j$-th class.
We initialize the label embedding following the conventional data mixing paradigm.
For example, when using CutMix to mix two images $\mathbf{X}_1$ and $\mathbf{X}_2$ from the $j$-th class and the $k$-th class with a mixing ratio of $\lambda$, we set $\tilde{y}_{cls, j} = \lambda$ and $\tilde{y}_{cls, k} = 1-\lambda$ for the class token.
For each patch token, we set $\tilde{y}_{i,j} =1$ if it comes from $\mathbf{X}_1$ and $\tilde{y}_{i,k} =1$ if it comes from $\mathbf{X}_2$.
If a patch token contains both the mixed images, we use the mixing ratio within this patch as the label.
For MixUp, we can simply set all label embeddings $\{ \mathbf{\tilde{y}}_i \}$ with $\tilde{y}_{,j} = \lambda$ and $\tilde{y}_{,j} = 1 - \lambda$.

We perform TL-Align in a layer-wise manner and compute the aligned labels based on the operation on the tokens.
Formally, ViTs use self-attention to perform spatial mixing of the input tokens $\mathbf{Z}$:
\begin{equation}
\begin{aligned}
& \mathbf{Q}=\mathbf{Z} \cdot \mathbf{W_{Q}} ,  \mathbf{K} = \mathbf{Z} \cdot \mathbf{W_{K}}, \mathbf{V}= \mathbf{Z} \cdot \mathbf{W_{V}},\\
& \mathcal{A}(\mathbf{Q},\mathbf{K}) = \text{Softmax}(\mathbf{Q} \cdot \mathbf{K}^{T} / \sqrt d),	\\
& \hat{\mathbf{Z}}=\text{SA}(\mathbf{Z})=\mathcal{A}(\mathbf{Q},\mathbf{K}) \cdot \mathbf{V}. \\
\end{aligned}
\end{equation}

To align the labels, we update the label embeddings $\mathbf{Y}$ using the same attention matrix $\mathcal{A}(\mathbf{Q},\mathbf{K})$:
\begin{equation} \label{eq:spatial_align}
\hat{\mathbf{Y}} =  \mathcal{A}(\mathbf{Q},\mathbf{K}) \cdot \mathbf{Y}.
\end{equation}
ViTs usually adopt multi-head self-attention (MSA) to perform multiple self-attentions parallelly:
\begin{equation} \label{eq:msa}
\hat{\mathbf{Z}} = \text{MSA}(\mathbf{Z}) = [\text{SA}_1(\mathbf{Z}); \text{SA}_2(\mathbf{Z}); \cdots; \text{SA}_H(\mathbf{Z})] \cdot \mathbf{w}_h,
\end{equation}
where $H$ is the number of heads and $\mathbf{w}_h \in \mathcal{R}^{d \times d}$.
We then adapt our label alignment to MSA by simply taking the average of all the attention matrices for alignment:
\begin{equation} \label{eq:msa_alignment}
\hat{\mathbf{Y}} = \text{TL-Align-S}(\mathbf{Z},\mathbf{Y}) \coloneqq \frac{1}{H} \sum_{i=1}^{H} \mathcal{A}_i(\mathbf{Q},\mathbf{K}) \cdot \mathbf{Y},
\end{equation}
where $\mathcal{A}_i$ is the attention matrix corresponding to the $i$-th head $\text{SA}_i$.

Each transformer block $l$ processes the tokens by both spatial and channel mixing:
\begin{equation} \label{eq:vit_block}
\begin{aligned}
& \hat{\mathbf{Z}}^{l-1} = \text{MSA}(\text{LN}(\mathbf{Z}^{l-1})),\ \  \mathbf{Z}'^{l-1} = \hat{\mathbf{Z}}^{l-1} +\mathbf{Z}^{l-1},	\\
& \hat{\mathbf{Z}}^l  = \text{MLP}(\text{LN}(\mathbf{Z}'^{l-1})), \ \ \mathbf{Z}^l = \hat{\mathbf{Z}}^l + \mathbf{Z}'^{l-1}, \\
\end{aligned}
\end{equation}
where MLP and LN denote the MLP module and layer normalization~\cite{ba2016layer}, respectively.
Our TL-Align then aligns the label embeddings in a similar manner:
\begin{equation} \label{eq:vit_align}
\begin{aligned}
& \hat{\mathbf{Y}}^{l-1} = \text{TL-Align-S}(\mathbf{Y}^{l-1}), \mathbf{Y}'^{l-1} = \text{Norm}(\hat{\mathbf{Y}}^{l-1} +\mathbf{Y}^{l-1}),	\\
& \hat{\mathbf{Y}}^l  = \mathbf{Y}'^{l-1}, \ \ \mathbf{Y}^l = \text{Norm}(\hat{\mathbf{Y}}^l + \mathbf{Y}'^{l-1}), \\
\end{aligned}
\end{equation}
where Norm denotes the normalization operation. We implement Norm by a simple average.

Hierarchical vision transformers such as Swin~\cite{liu2021swin} further introduce a patch aggregation operation to merge multiple patches.  
They usually concatenate multiple tokens across the channels to reduce the spatial resolution.
Instead of concatenation, we simply add the label embeddings of the merged tokens followed by normalization as the aligned labels. The proposed TL-Align can be generalized to different architectures composed of spatial mixing, channel mixing, point-wise transformation, residual connection, and spatial aggregation. We provide detailed illustrations of alignment with these operations in the appendix.

We synchronously align the labels with the processed tokens layer by layer and obtain the aligned tokens $\mathbf{Z}^L$ and labels $\mathbf{Y}^L$.
The final representation of the image $\mathbf{z}$ is either the class token $\mathbf{z}^L_{cls}$~\cite{dosovitskiy2020image, touvron2021training} or the average pooling of all the spatial tokens $\frac{1}{N} \sum_{i=1}^{N} \mathbf{z}^L_{i}$~\cite{liu2021swin}.
The aligned label $\mathbf{y}_{align}$ for the image is then $\mathbf{y}^L_{cls}$ or $\frac{1}{N} \sum_{i=1}^{N} \mathbf{y}^L_{i}$ depending on the specific model.
We then adopt the aligned label $\mathbf{y}_{align}$ to train the network and can adapt to different loss functions and training schemes:
\begin{equation} \label{eq:loss}
J = J (\mathbf{z}, \text{stop-gradient}(\mathbf{y}_{align})).
\end{equation}
We do not back-propagate through the aligned label as they only serve as a more accurate target.

Our TL-Align serves as a plug-and-play module on various vision transformers while only introducing negligible training costs. 
We adjust the label of each token adaptively during the layer-by-layer propagation and preserve alignment between tokens and labels throughout the forward process. 
TL-Align is only used during training and introduces no additional computation cost when inference.

%% file: chapters/4_experiment.tex
\section{Experiments} \label{experiment}
\newcommand{\tablesize}{\small}

In this section, we conducted extensive experiments to evaluate the proposed TL-Align method. 
We demonstrate the improvement of TL-Align on various vision transformers and compare it with state-of-the-art training strategies concerning accuracy, network complexity, and training speed. 
We examine the transferability on downstream tasks including semantic segmentation, object detection, and transfer learning. 
We further provide in-depth analysis to evaluate the effectiveness of TL-Align.

\begin{table}[t] \tablesize
  \centering
    \setlength{\tabcolsep}{0.015\linewidth}
\caption{\textbf{Results on ImageNet classification task.} We compare the parameters, FLOPs, and accuracy of different vision transformer backbones without and with our TL-Align.} 
\vspace{-3mm}
    \begin{tabular}{lccccc}\toprule
    Model & Image Size & Params & FLOPs  & Top-1(\%)  & Top-5(\%)  \\ \midrule
	DeiT-T   		& \multirow{2}{*}{$224^{2}$}   &   \multirow{2}{*}{5.7M} &  \multirow{2}{*}{1.6G}   &  72.2 & 91.3 \\   
	+TL-Align   									&  &    &    &  \textbf{73.2} & \textbf{91.7} \\ 
 PVT-T   				& \multirow{2}{*}{$224^{2}$} &   \multirow{2}{*}{13.2M} &  \multirow{2}{*}{1.9G}  &  75.1  & 92.4 \\  
	+TL-Align   									&   &    &    &  \textbf{75.5}  & \textbf{93.0}\\ \midrule 
	DeiT-S   		& \multirow{2}{*}{$224^{2}$}  &   \multirow{2}{*}{22M} 	&  \multirow{2}{*}{4.6G}  &  79.8 & 95.0 \\   
	+TL-Align  				 					&  &    &    &  \textbf{80.6} & 95.0 \\ 

 	PVT-S  				& \multirow{2}{*}{$224^{2}$}  &   \multirow{2}{*}{24.5M} &  \multirow{2}{*}{3.8G}  &  79.8 &  95.0 \\  
 	+TL-Align   									&   &    &    &  \textbf{80.4}  & \textbf{95.5}\\ 
	Swin-T   				& \multirow{2}{*}{$224^{2}$} &   \multirow{2}{*}{29M} &  \multirow{2}{*}{4.5G}  &  81.2  & 95.5\\  
	+TL-Align   									&   &    &    &  \textbf{81.4}  & \textbf{95.7}\\ \midrule
	Swin-S  				& \multirow{2}{*}{$224^{2}$}  &   \multirow{2}{*}{50M} &  \multirow{2}{*}{8.8G}  &  83.0 & 96.3 \\   
	+TL-Align   									&  &    &   &  \textbf{83.4} & \textbf{96.5} \\ 
  
	DeiT-B   		& \multirow{2}{*}{$224^{2}$}  &   \multirow{2}{*}{86M} &  \multirow{2}{*}{17.5G}  &  81.8 & 95.5 \\   
	+TL-Align   									&   &     &     &  \textbf{82.3} & \textbf{95.8} \\

	Swin-B   				& \multirow{2}{*}{$224^{2}$}  &   \multirow{2}{*}{88M} &  \multirow{2}{*}{15.4G}  &  83.5 & 96.4 \\   
	+TL-Align   									&   &    &    &  \textbf{83.7} & \textbf{96.5}\\
	     \bottomrule
    \end{tabular}%
  \label{tab:main_imagenet} 
  \vspace{-3mm}
\end{table}%

\begin{table}[t] \tablesize
  \centering
    \setlength{\tabcolsep}{0.015\linewidth}
\caption{\textbf{Comparison of our TL-Align with other training strategies on ImageNet.} }
\vspace{-3mm}
    \begin{tabular}{lcccc}\toprule
     Method & Params  & Speed (image/s) & Acc.(\%)\\ 
    \midrule
    Vanilla  					& 22M  	& 322  	& 76.4  \\
    \midrule
    CutMix 			& 22M  	& 322   & 79.8  \\
    Puzzle-Mix			& 22M  	& 139   & 79.8  \\
    SaliencyMix		& 22M  	& 314   & 79.2  \\
    Attentive-CutMix	 	& 46M  	& 239   & 77.5  \\
    TransMix			& 22M	& 322	&	80.1 \\
    \midrule
    CutMix	+ TL-Align	 				& 22M  & 311   & \textbf{80.6}  \\
	     \bottomrule
    \end{tabular}%
  \label{tab:comp_imagenet} 
  \vspace{-3mm}
\end{table}%

\begin{table}[t] \tablesize
  \centering
    \setlength{\tabcolsep}{0.015\linewidth}
\caption{\textbf{Results on semantic segmentation ADE20K.} }
\vspace{-3mm}
    \begin{tabular}{lccccc}\toprule
    Backbone &	Params	&	FLOPs & mIoU 	&	mIoU (MS) & mAcc \\\midrule
    DeiT-S	& \multirow{2}{*}{58M } & \multirow{2}{*}{1032G } & 43.8   & 45.1 & 55.2     \\
    +TL-Align 	&  &  &  \textbf{44.5} 	& \textbf{45.7} & \textbf{55.5}    \\\midrule
    
    Swin-T 	& \multirow{2}{*}{60M} & \multirow{2}{*}{945G}   & 44.4 & 45.8 & 55.6      \\
    +TL-Align 						&  &  &  \textbf{44.7} 	& \textbf{46.5} & \textbf{56.4}      \\\midrule
    
    Swin-S 	& \multirow{2}{*}{81M} & \multirow{2}{*}{1038G}   & 47.6 & 49.5& 58.8      \\
    +TL-Align 	&   &   & \textbf{48.0}  	& \textbf{49.7} & \textbf{59.5}   \\\midrule
    
    Swin-B 	& \multirow{2}{*}{121M} & \multirow{2}{*}{1188G} & 48.1   & 49.7 & 59.1       \\
    +TL-Align 	&  &  &   \textbf{48.3}	& \textbf{50.1} & \textbf{59.7}     \\\bottomrule
    \end{tabular}%
  \label{tab:segmentation} 
  \vspace{-6mm}
\end{table}%

\subsection{ImageNet Classification}
\textbf{Implementation Details.} 
We first evaluate TL-Align on ImageNet~\cite{russakovsky2015imagenet} for image classification.
ImageNet~\cite{russakovsky2015imagenet} contains $\sim$ 1.2M training images and 50K validation images from 1K categories and is a widely-used benchmark for performance evaluation. 
We implement our method based on PyTorch~\cite{paszke2019pytorch} and the timm library~\cite{rw2019timm}. 
We conduct experiments on various transformer architectures: three variants of DeiT~\cite{touvron2021training} (DeiT-T, DeiT-S, and DeiT-B), two variants of PVT~\cite{wang2021pyramid} (PVT-T, PVT-S) and three variants of Swin Transformer~\cite{liu2021swin} (Swin-T, Swin-S, and Swin-B). 
For tiny and small models, we train from scratch for 300 epochs following the same training recipe using CutMix as DeiT~\cite{touvron2021training}, PVT~\cite{wang2021pyramid} and Swin~\cite{liu2021swin}.
We keep all the data augmentation policies and all hyperparameter settings unchanged for fair comparisons. 
The only modification is that we replace the mixing targets in CutMiX with the labels obtained by our TL-Align.
For base models (i.e., Deit-B and Swin-B), we finetune the official pre-trained models for 40 epochs with a constant learning rate of 1e-5 and a weight decay of 1e-8.

 \begin{table*}[t]\tablesize
  \centering
  \caption{\textbf{Experimental results on object detection and instance segmentation on COCO.} } 
 \vspace{-3mm}
\setlength\tabcolsep{1 pt}
    \begin{tabu}to 0.8\textwidth{l|*{3}{X[c]}|*{3}{X[c]}|*{3}{X[c]}}\toprule 
    Backbone &	Params	&	FLOPs & Schedule 	&	$\text{AP}^{\text{box}}$		&	$\text{AP}_{\text{50}}^{\text{box}}$	& $\text{AP}_{\text{75}}^{\text{box}}$ &	$\text{AP}^{\text{mask}}$		&	$\text{AP}_{\text{50}}^{\text{mask}}$ & $\text{AP}_{\text{75}}^{\text{mask}}$	\\\midrule
    
    Swin-T 	& \multirow{2}{*}{86M} & \multirow{2}{*}{745G} & \multirow{2}{*}{3x}   & 50.4 & 69.2   & 54.7   & 43.7 & 66.6   & 47.3    \\
    +TL-Align 						&  &  &   	& \textbf{50.5} & \textbf{69.4}   & \textbf{54.9}   & \textbf{43.8} & 66.6   & 47.3    \\\midrule
    
    Swin-S 	& \multirow{2}{*}{107M} & \multirow{2}{*}{838G} & \multirow{2}{*}{3x}   & 51.9 & 70.7   & 56.3   & 45.0 & 68.2   & 48.8    \\
    +TL-Align 	&   &   &   	 &   \textbf{52.2}	& \textbf{71.1} &	\textbf{56.7} &	\textbf{45.2}	 & \textbf{68.4} &	\textbf{49.1}  \\\midrule
    
    Swin-B 	& \multirow{2}{*}{145M} & \multirow{2}{*}{982G} & \multirow{2}{*}{3x}   & 51.9 & 70.5   & 56.4   & 45.0 & 68.1   & 48.9    \\
    +TL-Align 	&  &  &   	& \textbf{52.3}	&\textbf{71.2} &	\textbf{56.9}	 &\textbf{45.3}	& \textbf{68.7} & 	\textbf{49.1}    \\\bottomrule

\end{tabu}%
\vspace{-5mm}
  \label{tab:detection}%
\end{table*}%

\begin{table}[t] \tablesize
  \centering
    \setlength{\tabcolsep}{0.015\linewidth}
\caption{\textbf{The accuracy and model complexity on different transfer learning datasets}. }
\vspace{-3mm}
    \begin{tabular}{lcccccc}\toprule
    Model   & Params  & FLOPs & C-10 & C-100 & Flowers & Cars \\\midrule
    ResNet50 					& 26M 	& 4.1G   & - & - & 96.2  & 90.0 \\
    ViT-B/16 			& 86M  & 55.4G   & 98.1  & 87.1  & 89.5  & - \\
    ViT-L/16			& 307M & 190.7G     & 97.9  & 86.4  & 89.7  & - \\
	\midrule
    
        Deit-T 	& 5.7M  & 1.6G    & 97.6  & 85.7  & 97.1  & 90.1 \\
       +TL-Align 							& 5.7M  & 1.6G    & \textbf{97.8}  & \textbf{86.4}  & \textbf{97.9}  & \textbf{90.7} \\
        Deit-S & 22M  & 4.6G     	& 97.9  & 90.2  & 98.1  & 91.4 \\
        +TL-Align 							& 22M  & 4.6G    	& \textbf{98.8}  & \textbf{90.4}  & \textbf{98.3}  & \textbf{91.8} \\
        Deit-B 	& 86M  & 17.5G   	& 99.1  & 90.8  & 98.4  & 92.1 \\
        +TL-Align 							& 86M  & 17.5G    & 99.1  & 90.5  & \textbf{98.6}  & \textbf{93.0} \\\bottomrule
    \end{tabular}%
  \label{tab:transfer} 
  \vspace{-5mm}
\end{table}%

\textbf{Performance on Different Architectures.} 
As shown in \cref{tab:main_imagenet}, TL-Align steadily improves the performance of different vision transformer architectures. 
 Specifically, TL-Align boosts the top-1 accuracy of DeiT-T, DeiT-S, and DeiT-B by 1.0\%, 0.8\%, and 0.5\%, respectively, in a parameter-free manner. 
Moreover, our method is generalizable and can be directly applied to hierarchical vision transformers like Swin. 
It is worth noting that most existing methods need either architecture modifications (adding a class token in ~\cite{chen2021transmix}) or extra computations (saliency map extraction in~\cite{uddin2020saliencymix}) when applied to Swin. 
In contrast, our TL-Align method can be used as a plug-and-play module and achieves consistent improvement on variants of Swin.

\begin{figure}[t]
	\centering
	\includegraphics[width=0.47\textwidth]{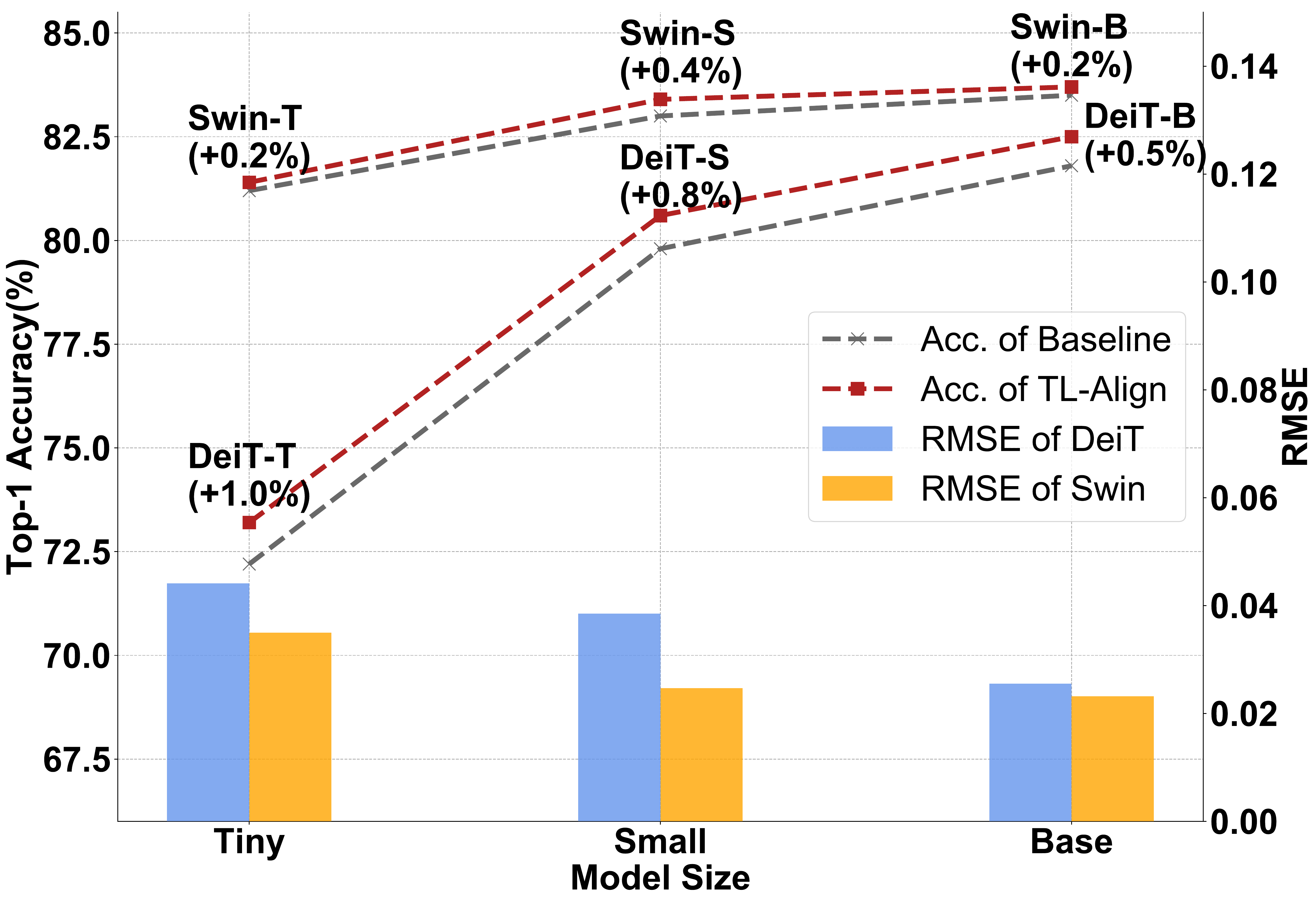}
\vspace{-3mm}
  \caption{\textbf{The Root Mean Square Error (RMSE) between original CutMix targets and labels obtained by T-L Align.} We show results on variants of DeiT and Swin.
	  } 
  \label{fig:rmse}
 \vspace{-7mm}
 \end{figure}

\begin{figure}[t]
	\centering
	\includegraphics[width=0.47\textwidth]{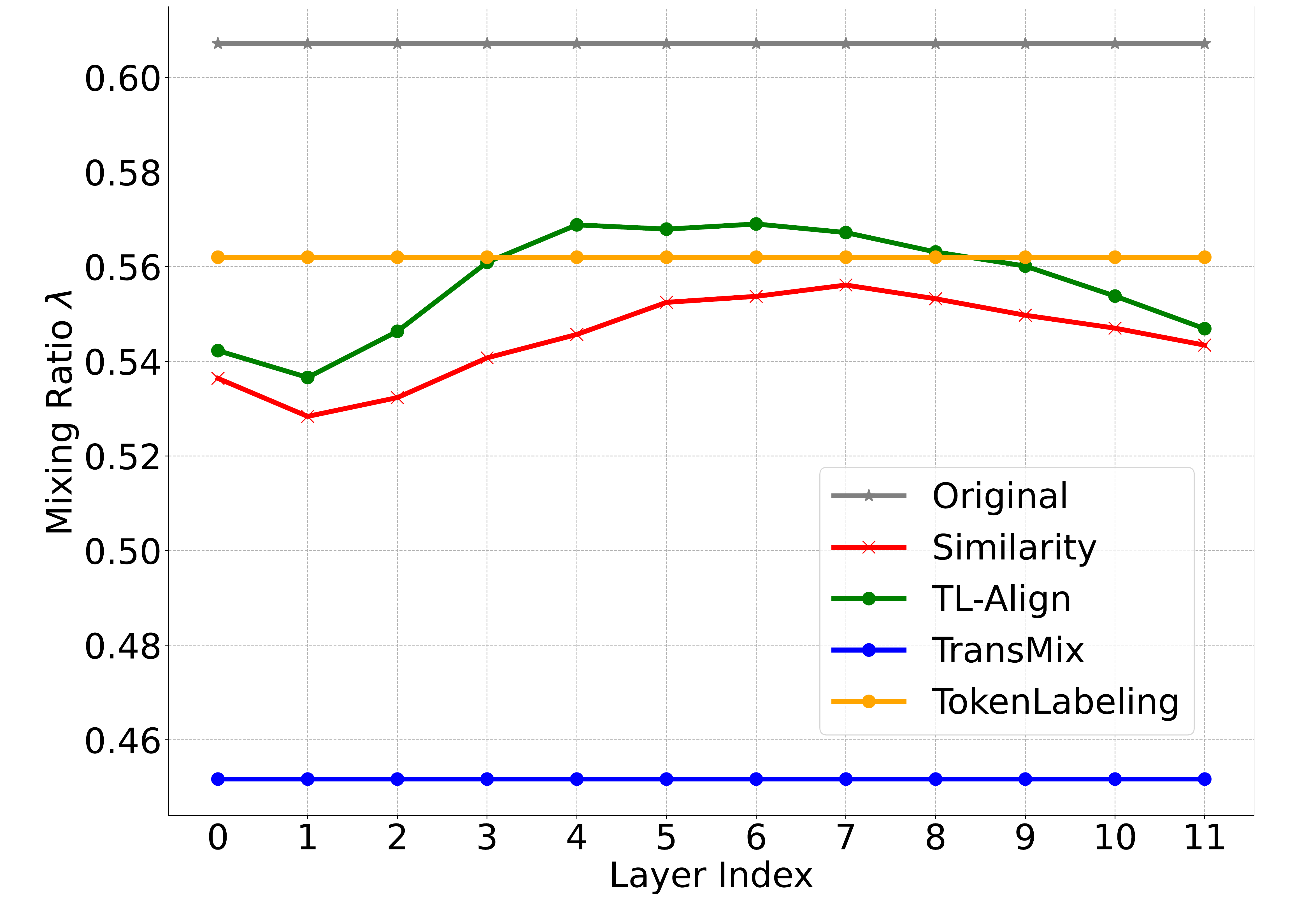}
\vspace{-3mm}
  \caption{\textbf{Visualization of mixing ratio $\lambda$ of fluctuating tokens from different layers.} We compare the results of TL-Align with CutMix, token similarity, TransMix, and TokenLabeling.
	  } 
  \label{fig:ratio}
 \vspace{-7mm}
 \end{figure} 

\begin{table*}[t] \tablesize
  \centering
    \setlength{\tabcolsep}{0.015\linewidth}
\caption{\textbf{Comparison results of model generalization ability and robustness.} 
  We evaluate them on various out-of-distribution/corrupted datasets and against adversarial attacks. 
  $\uparrow$ denotes higher is better, and $\downarrow$ denotes lower is better. }
\vspace{-3mm}
    \begin{tabular}{lccccccccc}\toprule
\multicolumn{1}{l}{\multirow{2}[0]{*}{Model}} & \multirow{2}[0]{*}{FLOPs} & \multirow{2}[0]{*}{Params} & \multicolumn{2}{c}{ImageNet} & \multicolumn{1}{c}{Generalization} & \multicolumn{4}{c}{Robustness} \\
        
        \cmidrule(lr){4-5} \cmidrule(lr){6-6} \cmidrule(lr){7-10}
          & & & \multicolumn{1}{c}{Top-1$\uparrow$} & \multicolumn{1}{c}{Top-5$\uparrow$} & \multicolumn{1}{c}{IN-V2$\uparrow$} & \multicolumn{1}{c}{IN-A$\uparrow$} & \multicolumn{1}{c}{IN-C$\downarrow$} & \multicolumn{1}{c}{IN-R$\uparrow$} & \multicolumn{1}{c}{AutoAttack$\uparrow$}\\\midrule
    DeiT-T 					& 5.7M 	& 1.6G  & 72.2 	&  91.3  &  60.4  &  7.7  &  69.1  & 34.1  &  3.9 \\
    +TL-Align 		& 5.7M 	& 1.6G	& \textbf{73.2}	&  \textbf{91.7}  &  \textbf{61.4}  &  6.1  &  \textbf{68.0}  & \textbf{34.6}  &  \textbf{4.4} \\\midrule
    DeiT-S 					& 22M   & 4.6G 	& 79.8  &  95.0  &  68.5  &  18.9  &  54.7 &  42.5  &  6.9  \\
    +TL-Align 		& 22M   & 4.6G  & \textbf{80.6}  &  95.0  &  \textbf{68.9}  &  \textbf{19.2}  &  \textbf{53.2} &  \textbf{43.2}  &  \textbf{7.5} \\\midrule
    DeiT-B 					& 86M 	& 17.5G & 81.8  &  95.5  &  70.5  &  27.9  &  48.5 &  45.3   & -\\
    +TL-Align 		& 86M 	& 17.5G	& \textbf{82.3}  &  \textbf{95.8}  &  \textbf{70.9}  &  \textbf{29.0}  &  \textbf{47.1} & 44.4 & - \\
        \bottomrule
    \end{tabular}%
  \label{tab:robust} 
  \vspace{-5mm}
\end{table*}%

\begin{table}[t] \tablesize
  \centering
    \setlength{\tabcolsep}{0.015\linewidth}
\caption{\textbf{Ablation of applying TL-Align to different data mixing strategies for DeiT-S training.} }
\vspace{-3mm}
    \begin{tabular}{cccc|cc}\toprule
    \multirow{2}[0]{*}{MixUp}  & \multirow{2}[0]{*}{CutMix} & \multirow{2}[0]{*}{Random} & \multirow{2}[0]{*}{Block-wise} & \multirow{2}[0]{*}{Top-1 (\%)} & +TL-Align \\
       &  &  &  &  & Top-1 (\%) \\\midrule
     $\times$ 	& $\times$   & $\times$ &  $\times$  & 76.4 & - \\
     $\times$    & $\checkmark$   & $\times$  & $\times$  & 79.8  & 80.6 \\
     $\checkmark$  & $\checkmark$   & $\times$  & $\times$   & 79.8  & 80.2 \\
     $\times$ & $\times$  & $\checkmark$   & $\times$  & 79.7  & 80.2   \\
     $\times$ & $\times$  & $\times$   & $\checkmark$  & 80.0  & 80.3   \\
    \bottomrule
    \end{tabular}%
  \label{tab:datamix} 
  \vspace{-6mm}
\end{table}%

\textbf{Comparison with Other Training Strategies.}
We also compare our method with the state-of-the-art training strategies for data mixing on DeiT-S, including CutMix~\cite{yun2019cutmix}, Puzzle-Mix~\cite{kim2020puzzle}, SaliencyMix~\cite{uddin2020saliencymix}, Attentive-CutMix~\cite{walawalkar2020attentive}, and TransMix~\cite{chen2021transmix}. 
Specifically, we train the DeiT-S model while only disabling CutMix as the baseline method, which is denoted as Vanilla in \cref{tab:comp_imagenet}. 
Moreover, since TransMix~\cite{chen2021transmix} reports the EMA accuracy with different hyperparameters, we reproduce it under the same training recipe \cite{touvron2021training} for a fair comparison.
 As demonstrated in \cref{tab:comp_imagenet}, TL-Align shows significantly better performance than the other mixup variants while maintaining the number of parameters and training speed.
 Puzzle-Mix obtains the same classification accuracy as CutMix but results in a much lower training speed as it relies on an extra model to get the optimal solution. 
 SaliencyMix and Attentive-CutMix lead to performance degeneration when built upon DeiT-S backbone.
 Notably, our method also achieves higher top-1 accuracy than ViT-targeted TransMix. 
 Due to the token fluctuation phenomenon, the class token attention utilization in TransMix can not reflect the actual contribution of different tokens. 
Differently, TL-Align obtains accurate alignment of the tokens and labels, resulting in improved performance.

\subsection{Downstream Tasks}
\textbf{Semantic Segmentation.}
We evaluate our TL-Align on ADE20K dataset~\cite{zhou2019semantic} for semantic segmentation. 
ADE20K~\cite{zhou2019semantic} contains 20K training images and 2K validation images from 150 semantic categories. 
We adopt DeiT-S and three variants of Swin Transformer as backbones equipped with UpperNet for segmentation. 
As presented in \cref{tab:segmentation}, TL-Align improves the segmentation performance on both DeiT and Swin at different model scales.

\textbf{Object Detection and Instance Segmentation.}
We also examine the performance of TL-Align on object detection and instance segmentation on the COCO 2017 dataset~\cite{lin2014microsoft}, which consists of 118K training images and 5K validation images from 80 categories. 
We apply our TL-Align to Swin~\cite{liu2021swin} due to the advantage of the hierarchical representations on object detection tasks. 
We adopt the Cascade Mask-RCNN~\cite{cai2018cascade} framework and use the training strategy of 3x schedule. 
As shown in \cref{tab:detection}, we observe consistent improvements on all variants of Swin Transformer. 
This demonstrates the advantages to learn token-level meaningful features suitable for dense prediction tasks.

\textbf{Transfer Learning.}
We further evaluate the transferred classification performance of TL-Align on CIFAR-10~\cite{krizhevsky2009learning}, CIFAR-100~\cite{krizhevsky2009learning}, Flowers~\cite{nilsback2008automated} and Cars~\cite{krause20133d}. 
We use pre-trained models on ImageNet and finetune them on these datasets following existing works~\cite{touvron2021training}. 
We compare the performance with and without TL-Align on three variants of DeiT~\cite{touvron2021training}, as shown in \cref{tab:transfer}. 
TL-Align obtains significant performance gains for all variants on the four datasets.

\subsection{Performance Analysis and Visualization}

\textbf{Effectiveness of Token-Label Alignment.} 
We first quantize the difference between the original targets and aligned labels and investigate its correlation with the model.
Specifically, we compute the Root Mean Square Error (RMSE) between the original targets and labels obtained by our TL-Align. 
As shown in \Cref{fig:rmse}, the RMSE decreases when enlarging the model size.
This indicates that larger models demonstrates less token fluctuation since the self-correction ability is also enhanced as the model capacity scales up.
Moreover, the RMSE for Swin Transformer tends to be lower compared with DeiT of similar model size. 
This is due to the local-window self-attention in Swin which preserves more local information.
These observations are consistent with our experimental results: the improvements on small models and DeiT-like backbones tend to be more significant as they encounter more token fluctuation.

\textbf{Visualization of the Layer-wise Mixing Ratio of Fluctuated Tokens.}
To investigate the effectiveness of TL-Align, we compute a similarity-based ``ground-truth" mixing ratio for each layer.
Specifically, we compute the similarities of tokens between the mixed and unmixed images and use them as the label of each token. 
We compare them with the mixing ratios produced by TL-Align, CutMix~\cite{yun2019cutmix}, TransMix~\cite{chen2021transmix}, and TokenLabeling~\cite{jiang2021all}.
As shown in~\Cref{fig:ratio}, the similarity-based mixing ratio changes at each layer, resulting from token fluctuation. 
However, CutMix, TransMix, and TokenLabeling assume the output tokens keep spatial correspondence with the input tokens and compute a fixed mixing ratio. 
TL-Align assigns dynamic labels to tokens using layer-wise alignment, which is more accurate compared with other methods.

\begin{table}[t] \tablesize
  \centering
    \setlength{\tabcolsep}{0.015\linewidth}
\caption{\textbf{Ablation of different TL-Align operations.} }
\vspace{-3mm}
    \begin{tabular}{l|cc}\toprule
     Alignment &Top-1 Acc.(\%) & $\Delta$ (\%) \\ \midrule
       None (DeiT-S baseline)& 79.8 & - \\
      TL-Align-S (Layer 12) & 80.1 & +0.3 \\
      TL-Align-S (Layer 2,4,6,8)& 80.2 & +0.4 \\
     Normalization Disabled&80.3 & +0.5\\
      Default (TL-Align)& \textbf{80.6} & \textbf{+0.8}\\
    \bottomrule
    \end{tabular}%
  \label{tab:ablation_align} 
  \vspace{-7mm}
\end{table}%

\begin{figure*}[t]
	\centering
	\includegraphics[width=\textwidth]{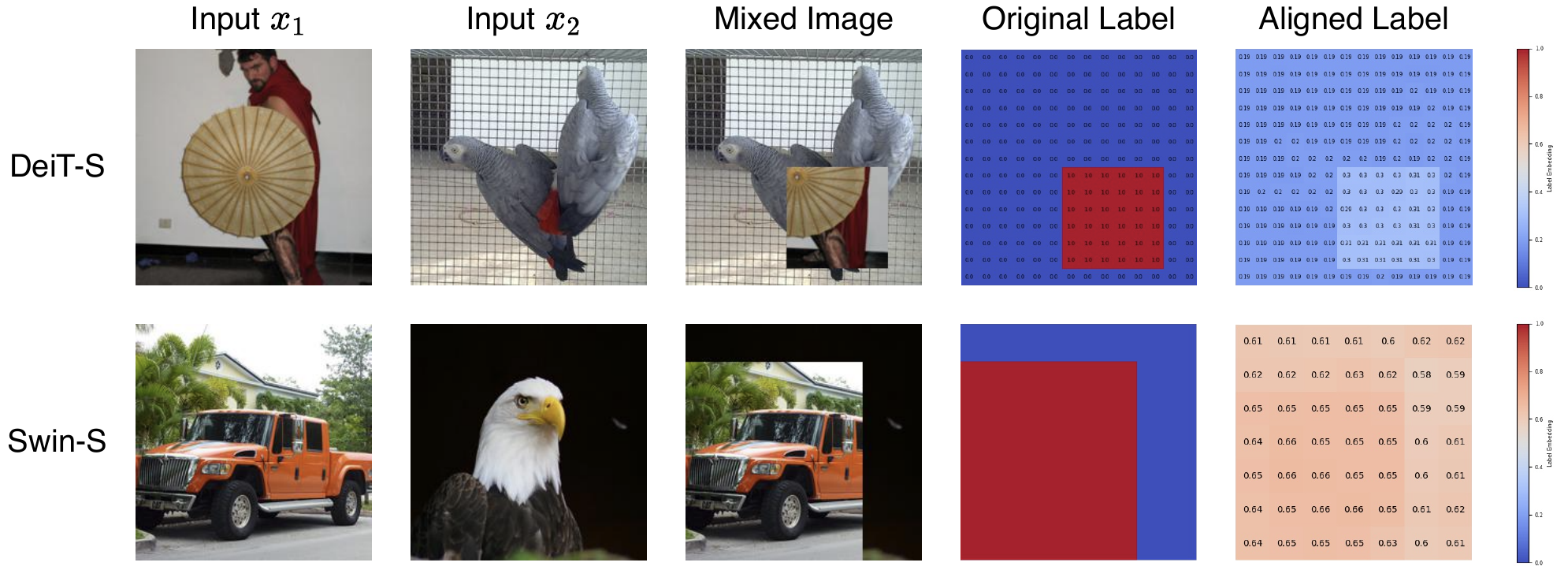}
\vspace{-8mm}
  \caption{
\textbf{The visualization results on DeiT-S and Swin-S.} We visualize the input images, the mixed image, the original label embedding, and the label embedding after token-label alignment. 	  } 
  \label{fig:vis_label}
 \vspace{-7mm}
 \end{figure*}  

\textbf{Evaluation of Robustness and Generalization.}
We further conduct experiments to validate the generalization and robustness of TL-Align, as shown in \cref{tab:robust}. 
We employ four corrupted and out-of-distribution datasets for robustness evaluation.
ImageNet-A\cite{hendrycks2021natural} consists of naturally adversarial examples from real-world challenging scenarios. 
ImageNet-C~\cite{hendrycks2019benchmarking} is used to evaluate model robustness to diverse image corruptions. 
ImageNet-R~\cite{hendrycks2021many} contains various artistic renditions of 200 ImageNet classes.
We also adopt AutoAttack~\cite{croce2020reliable} to evaluate the adversarial robustness on the ImageNet validation set. 
Due to memory limitation, we do not experiment with DeiT-B on AutoAttack.
We use mean Corruption Error (mCE) for ImageNet-C and Top-1 Accuracy for others as the evaluation metric. 
For generalization evaluation, we adopt the ImageNet-V2 dataset~\cite{recht2019imagenet} which contains new test sets of ImageNet following the same labeling protocol. 
We see that TL-Align improves both robustness and generalization, showing the superiority of adopting TL-Align for pre-training.

\textbf{Ablation Study on different Data Mixing Strategies.}
Due to the efficiency of the proposed layer-wise alignment, TL-Align can be directly applied to a wide range of data mixing strategies. We adopt MixUp, CutMix, a random mixing strategy and a block-wise mixing strategy to evaluate the generalizability of TL-Align. The random mixing and block-wise mixing strategies are inspired by MAE~\cite{he2022masked} and BEiT~\cite{bao2021beit} and we replace the masking operation with image mixing on patch-level and block-level (both of size 16$\times$16) respectively. The comparison results of training DeiT-S with and without our approach is demonstrated in~\cref{tab:datamix}. Specifically, TL-Align improves CutMix by 0.8\%, MixUp+CutMix by 0.4\%, random mixing by 0.5\% and block-wise mixing by 0.3\% respectively, further verifying the generalizability of the proposed TL-Align.

\textbf{Ablation Study on Different Label Alignment Operations.}
Our TL-Align aligns the labels with tokens transformed by spatial self-attention and residual connection layer-by-layer. To investigate the effect of reusing attention maps and normalization, we conduct an ablation study regarding different alignment operations on DeiT-S. We try aligning the labels only by using the attention map of Layer 12, which is equivalent to TransMix~\cite{chen2021transmix}. We also test the performance of applying alignment to several middle transformer layers and disabling normalization. As presented in \cref{tab:ablation_align}, incomplete alignment at a part of layers marginally boosts the performance as it cannot well handle the token fluctuation issue. Disabling normalization leads to 0.3$\%$ accuracy drop due to the inaccurate alignment at the presence of residual connections. This demonstrates the significance of the token-label alignment by attention utilization and normalization in a layer-wise manner.

\textbf{Visualizations of Aligned Labels.}
We visualize the labels obtained by TL-Align on DeiT-S~\cite{touvron2021training} and Swin-S~\cite{liu2021swin} as shown in \Cref{fig:vis_label}. 
 Specifically, the aligned label embedding is obtained after the final transformer block for both DeiT-S and Swin-S. 
 The value of the label embedding represents the probability of the belonged class of the corresponding token. 
We use red to denote larger probabilities towards the first image and blue for the second image.
We observe that the aligned labels can deviate from the original labels and result in different mixing ratios for training.
Therefore, using the original ratio as the training target may produce false training signals and lead to inferior performance.
We see that TL-Align can correct the labels when the images are mixed with uninformative tokens.
More visualization results are included in the appendix.

%% file: chapters/5_conclusion.tex
\section{Conclusion} \label{conclusion}
In this paper, we have presented a token-label alignment method for training better vision transformers.
As important subsets of data augmentation methods, data mixing strategies can generally improve the performance of both CNNs and ViTs.
We identify a token fading issue for ViTs and address it by tracing the correspondence between transformed tokens and the original tokens to obtain a label for each output token to obtain more accurate training signals.
Experimental results have demonstrated that our TL-Align can consistently improve the performance of various ViT models.
The generalization performance of TL-Align to other architectures such as MLP-like models remains unknown and is a promising future direction to explore.

%% file: chapters/a_training.tex
 \section{Comparisons of Different Training Recipes}
We compare different training recipes for the DeiT-S model in~\cref{tab:recipe}.
The results of TransMix~\cite{chen2021transmix} reported in the original paper adopts an advanced training recipe with a model exponential moving average, resulting in slower training speed.
Differently, we basically follow the conventional DeiT-S~\cite{touvron2021training} training recipe and improve its performance by 0.8\%.
We report the result of TransMix with the same training recipe (80.1\%) in Table~\ref{tab:comp_imagenet}.

\begin{table*}[t] \tablesize
  \centering
    \setlength{\tabcolsep}{0.009\linewidth}
\caption{\textbf{Comparisons of different training recipes for the DeiT-S model on ImageNet-1K.} }
\vspace{-3mm}
    \begin{tabular}{l|ccccccccccc}\toprule
            \multirow{2}[0]{*}{Method} & Training & Warmup & \multirow{2}[0]{*}{LR} & Weight  & Model  & EMA & \multirow{2}[0]{*}{MixUp} & \multirow{2}[0]{*}{CutMix}  & MixUp  & Random  & Top-1 \\
             & Epochs & Epochs & & Decay & EMA & Decay & & & Switch Prob & Erasing & Acc. (\%)\\
            
\midrule
DeiT-S$^1$~\cite{touvron2021training}& 300 & 5 & 0.0005 & 0.05 & $\times$ & -  & 0.0  & 0.0 & - & $\checkmark$ & 76.4 \\
DeiT-S$^2$~\cite{touvron2021training}& 300 & 5 & 0.0005 & 0.05 & $\times$ & -  & 0.8  & 1.0 & 0.5 & $\checkmark$ & 79.8 \\ \midrule
DeiT-S$^3$~\cite{touvron2021training}& 310 & 20 & 0.001 & 0.03 & $\checkmark$ & 0.99996  & 0.8  & 1.0 & 0.8 & $\times$ & 80.3 \\
+TransMix~\cite{chen2021transmix}& 310 & 20 & 0.001 & 0.03 & $\checkmark$ & 0.99996  & 0.8  & 1.0 & 0.8 & $\times$ & 80.7 (+0.4) \\ \midrule
DeiT-S$^4$~\cite{touvron2021training}& 300 & 5 & 0.0005 & 0.05 & $\times$ & -  & 0.0  & 1.0 & - & $\checkmark$ & 79.8 \\
+TransMix~\cite{chen2021transmix}& 300 & 5 & 0.0005 & 0.05 & $\times$ & -  & 0.0  & 1.0 & - & $\checkmark$ & 80.1 (+0.3) \\
+TL-Align & 300 & 5 & 0.0005 & 0.05 & $\times$ & -  & 0.0  & 1.0 & - & $\checkmark$ & \textbf{80.6 (+0.8)} \\

    \bottomrule
    \end{tabular}%
  \label{tab:recipe} 
\end{table*}%

%% file: chapters/a_exp_details.tex
\section{Details of Experimental Analysis} \label{exp_details}
\begin{figure*}[t]
	\centering
	\includegraphics[width=\textwidth]{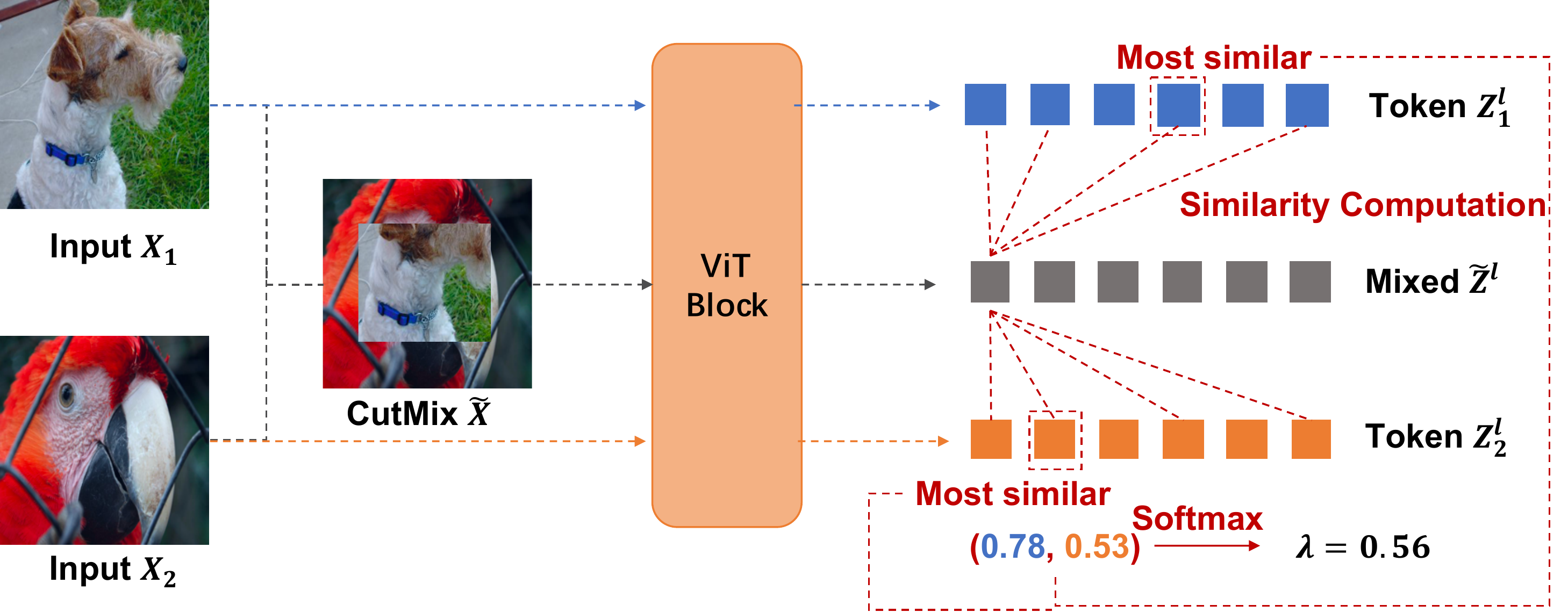}
\vspace{-5mm}
  \caption{
Illustration of how we get a ``ground-truth" mixing ratio based on token similarity.} 
  \label{fig:gt}
 \vspace{-5mm}
 \end{figure*}

\paragraph{Obtaining the ``Ground-truth''  Mixing Ratio.} 
To better demonstrate the token fluctuation phenomenon, we compute a ``ground-truth'' mixing ratio based on token similarity as shown in~\Cref{fig:gt}. 
Formally, given two input images $\mathbf{X}_1$, $\mathbf{X}_2$ and their mixed sample $\mathbf{\mathbf{\tilde X}}$ generated by CutMix, we feed all of them into the vision transformer obtain get the corresponding tokens $\mathbf{Z}^{l}_{1}$, $\mathbf{Z}^{l}_{2}$ and $\tilde {\mathbf{Z}}^{l}$ after the transformer block $l$. 
For each mixed token $\tilde {\mathbf{z}}^{l}_{i}$ in $\tilde {\mathbf{Z}}^{l}$, we compute its maximum cosine similarity with all tokens in $\mathbf{Z}^{l}_{1}$ and $\mathbf{Z}^{l}_{2}$: 
\begin{equation}
\begin{aligned}
\mathbf{s}^{l}_{1}(\tilde {\mathbf{z}}^{l}_{i})=\max_{j} \frac{ (\tilde {\mathbf{z}}^{l}_{i}) ^{T}  {\mathbf{z}}^{l}_{1j} }{||\tilde {\mathbf{z}}^{l}_{i}||\cdot|| {\mathbf{z}}^{l}_{1j}||}, \\
\mathbf{s}^{l}_{2}(\tilde {\mathbf{z}}^{l}_{i})=\max_{j} \frac{ (\tilde {\mathbf{z}}^{l}_{i}) ^{T}  {\mathbf{z}}^{l}_{2k} }{||\tilde {\mathbf{z}}^{l}_{i}||\cdot|| {\mathbf{z}}^{l}_{2k}||}. 	
\end{aligned}
\end{equation}
The contribution of input  $\mathbf{X}_1$ to the token $\tilde {\mathbf{z}}^{l}_{i}$ is then obtained using the softmax function: $\lambda = \text{softmax} (\mathbf{s}^{l}_{1}(\tilde {\mathbf{z}}^{l}_{i}),\mathbf{s}^{l}_{2}(\tilde {\mathbf{z}}^{l}_{i}))$. 
We visualize this similarity-based mixing ratio of the class token in DeiT-S in Figure~\ref{fig:ratio}. 
The token mixing ratio changes after each transformer block's processing, demonstrating the token fluctuation problem. 
Moreover, TL-Align assigns a dynamic mixing ratio to tokens at different layers, which is more consistent with the ``ground truth'' compared with other methods. 
This provides an empirical analysis to explain the improvement achieved by our TL-Align. 

\begin{figure*}[t]
	\centering
	\includegraphics[width=\textwidth]{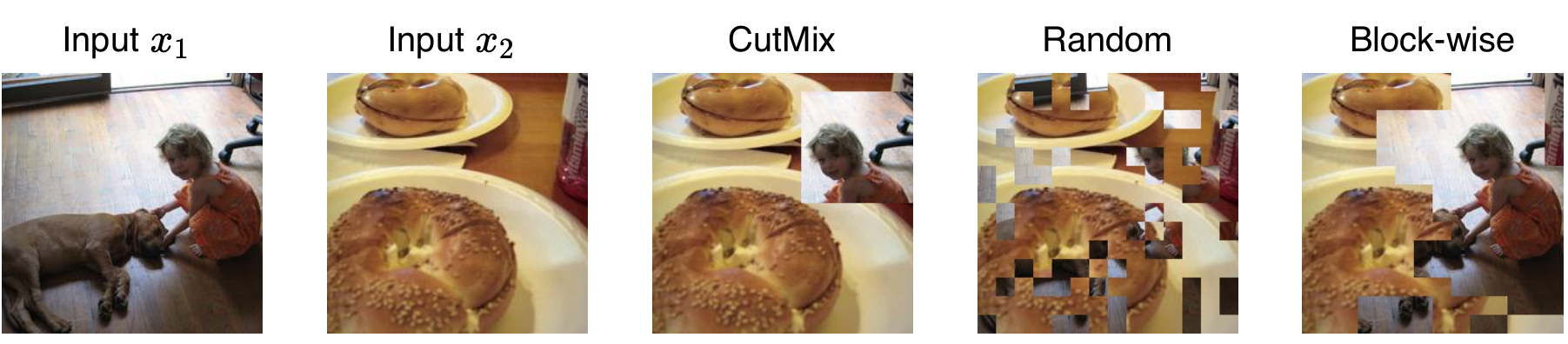}
\vspace{-5mm}
  \caption{
Visualization of mixed images produced by different data mixing strategies.} 
  \label{fig:mix}
 \end{figure*}

\paragraph{Implementation of Different Data Mixing Strategies.} 
We provide implementation details of different data mixing strategies that we adopt to evaluate the effectiveness of TL-Align. 
Inspired by MAE~\cite{he2022masked} and BEiT~\cite{bao2021beit}, we implement a random mixing strategy and block-wise mixing strategy. 
The visualization of the mixed images produced by CutMix, random mixing, and block-wise mixing strategies is shown in~\Cref{fig:mix}. 
Specifically, employing the block-wise strategy leads to the highest top-1 accuracy of $80.0\%$.
Our TL-Align further boosts the accuracy by $+0.3\%$, verifying its generalizability on various data mixing strategies.

%% file: chapters/a_vis.tex
\section{More Visualization Results} \label{appen:vis}

\begin{figure*}[t]
	\centering
	\includegraphics[width=\textwidth]{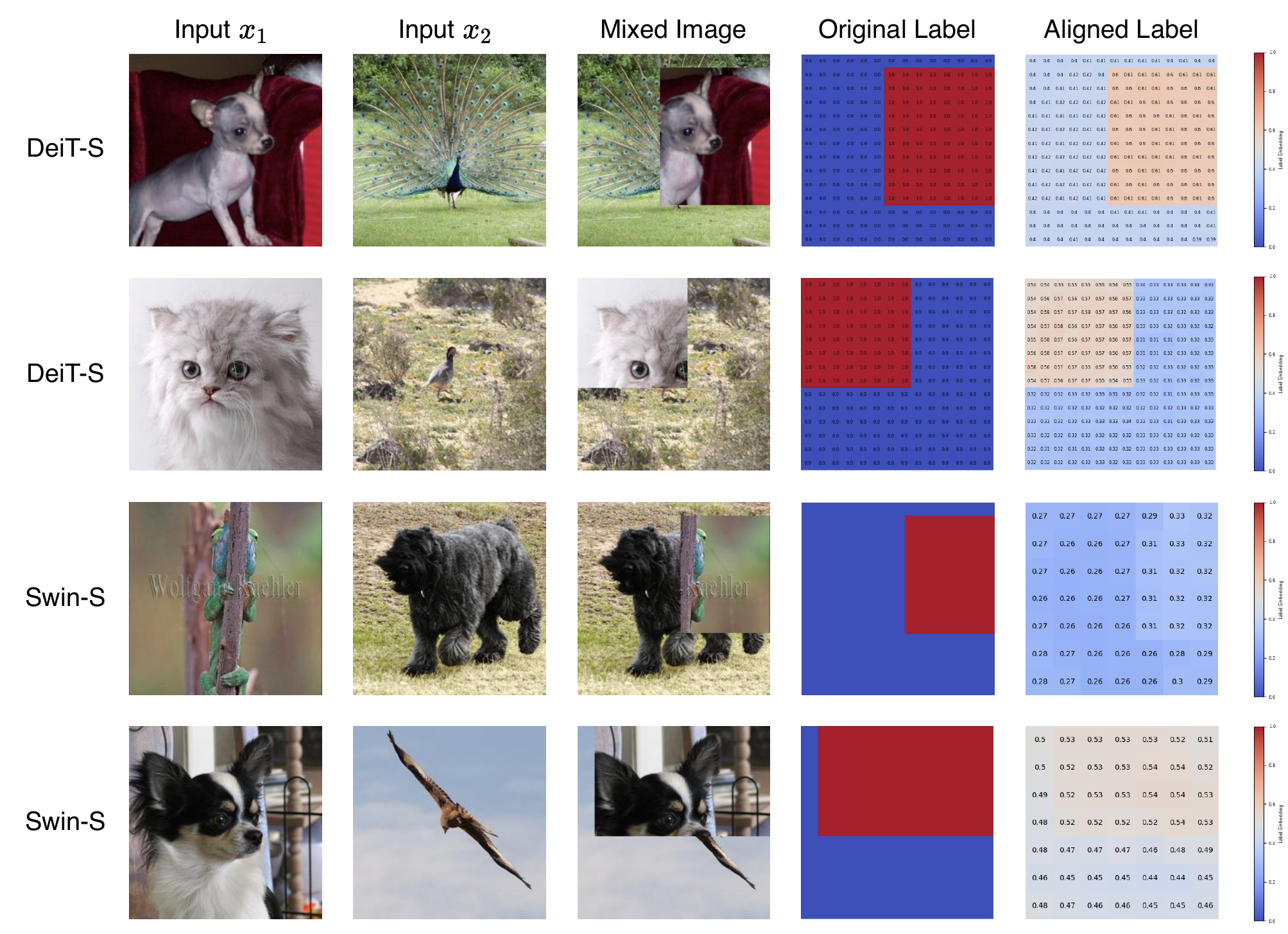}
\vspace{-7mm}
  \caption{
More visualization results on DeiT-S and Swin-S. We visualize the input images, the mixed image, the original label embedding, and the label embedding after token-label alignment.	  } 
  \label{fig:vis_supp}
 \end{figure*}

We provide more visualization results of the obtained labels by the proposed token-label alignment method in~\Cref{fig:vis_supp}. 
 We visualize the input images, the mixed image, the original label embedding, and the label embedding after our TL-Align. 
 Specifically, we visualize the aligned label embedding after the final transformer block for both DeiT-S and Swin-S. 
 The size of the original label embedding is equivalent to the number of input tokens, i.e., $14\times 14$ for DeiT-S and $56\times 56$ for Swin-Transformer since they employ different patch sizes for patch embedding. 
 The value of the label embedding represents the probability of which class the corresponding token belongs to, which is shown by color. 
Red stands for the class of the first input image while blue stands for the class of the second input image. 
We observe that the aligned labels can deviate from the original labels, resulting in different mixing ratios during training.
Therefore, using the original mixing ratio as the training target produces false training signals and might lead to inferior performance.

%% file: chapters/a_generalize.tex
 \section{Generalizing TL-Align Beyond ViTs}
ViTs can achieve better accuracy/computation trade-off than conventional CNNs, where one of the working mechanisms is the alternation between spatial mixing (e.g., SA) and channel mixing (e.g., MLP)~\cite{tolstikhin2021mlp}.
Based on this, some works have explored different spatial mixing strategies in addition to self-attention, including spatial MLP~\cite{tolstikhin2021mlp, touvron2021resmlp, tang2021image,wei2022activemlp} and depth-wise convolution~\cite{ding2022scaling,liu2022convnet,guo2022visual}.
For an image $\mathbf{X} \in \mathbb{R}^{H\times W \times C}$, they first perform patch-wise image tokenization to obtain a tokenized image representation $\mathbf{Z} \in \mathbb{R}^{N \times d} $, where $N$ is the number of tokens and $d$ is the number of channels.
To generalize TL-Align to other architectures beyond ViTs, we first formulate modern deep vision networks into various compositions of five operations:
\begin{itemize}
	\item Spatial mixing: $\mathbf{Z} \leftarrow \mathbf{W}^s(\mathbf{Z}) \cdot  \mathbf{Z} $, where $\mathbf{W}^s(\mathbf{Z}) \in \mathbb{R}^{N \times N} $.
	\item Channel mixing: $\mathbf{Z} \leftarrow \mathbf{Z} \cdot  \mathbf{W}^c(\mathbf{Z}) $, where $\mathbf{W}^c(\mathbf{Z}) \in \mathbb{R}^{d \times d} $.
	\item Point-wise transformation: $\mathbf{Z} \leftarrow f(\mathbf{Z})$, where $f$ is a point-wise operation such as bias adding and normalization.
	\item Residual connection: $\mathbf{Z} \leftarrow \mathbf{Z} + g(\mathbf{Z})$, where $g$ can be one or a composition of the aforementioned operations.
	\item Spatial aggregation: $\mathbf{Z} \leftarrow \text{Aggre}(\{\mathbf{Z}_i\}) $, where Aggre typically concatenates multiple tokens across the feature dimension.
\end{itemize}
For example, MLP-Mixer~\cite{tolstikhin2021mlp} adopts $\mathbf{W}^s(\mathbf{Z}) = W^s$, where $W^s \in \mathbb{R}^{N \times N}$ is a learnable parameter matrix.
ConvNeXt~\cite{liu2022convnet} adopts $\mathbf{W}^s(\mathbf{Z}) = T(\mathbf{K})$, where $\mathbf{K} \in \mathbb{R}^{7 \times 7}$ is a convolutional kernel and $T$ transforms the kernel into a equivalent matrix for direct multiplication.

The proposed TL-Align can be generalized to different architectures by applying the corresponding operations on the label embeddings.
We initialize the label embedding following \eqref{eq:input_y}.
We detail the label embedding updating for different operations in \cref{tab:tl-align}.
The $\text{Norm}(\cdot)$ operation denotes that we normalize each row vector so that the sum of all elements equals to 1.

\begin{table*}[!tb] \tablesize
  \centering
    \setlength{\tabcolsep}{0.015\linewidth}
\caption{\textbf{Updating of the label embeddings for different operations on the tokens.} }
\vspace{-2mm}
    \begin{tabular}{lccc}\toprule

       Operation   & Token Processing  & Label Alignment & Example \\\midrule
    Spatial mixing 	& $\mathbf{Z} \leftarrow \mathbf{W}^s(\mathbf{Z}) \cdot  \mathbf{Z} $ 	& $\mathbf{Y} \leftarrow \text{Norm} (\mathbf{W}^s(\mathbf{Z})) \cdot  \mathbf{Y} $ & Spatial attention \\
    Channel mixing 	& $\mathbf{Z} \leftarrow \mathbf{Z} \cdot  \mathbf{W}^c(\mathbf{Z}) $ 	& $\mathbf{Y} \leftarrow \mathbf{Y} $ & Channel MLP \\
    Point-wise transformation 	& $\mathbf{Z} \leftarrow f(\mathbf{Z})$ 	& $\mathbf{Y} \leftarrow \mathbf{Y} $ & Layer normalization \\
    Residual connection 	& $\mathbf{Z} \leftarrow \mathbf{Z} + g(\mathbf{Z})$ 	& $\mathbf{Y} \leftarrow \text{Norm} (\mathbf{Y} + g(\mathbf{Y}))$ & Residual connection \\
    Spatial aggregation 	& $\mathbf{Z} \leftarrow \text{Aggre}(\{\mathbf{Z}_i\}) $	& $\mathbf{Y} \leftarrow \text{Norm} (\sum_i \mathbf{Y}_i) $ & Patch merging\\
    \bottomrule
    \end{tabular}%
  \label{tab:tl-align} 
  \vspace{-5mm}
\end{table*}%

For spatial mixing, we accordingly mix the token embeddings using the same weights as the token processing.
For example, for a processed token $\mathbf{\hat{z}} = 0.5 \cdot \mathbf{z}_1 + 0.5 \cdot \mathbf{z}_2$, we similarly compute the aligned label as $\mathbf{\hat{y}} = 0.5 \cdot \mathbf{y}_1 + 0.5 \cdot \mathbf{y}_2$, assuming the label information is linearly addable.
As channel mixing and point-wise transformation only reorganize information within each token, they do not alter the label embedding.
For residual connection, we similarly add a residual connection to the label embedding before normalization.
Spatial aggregation is similar to spatial mixing and also aggregates information among multiple tokens.
Therefore, we also need to align the labels by adding their label embeddings before normalization.
We leave the experiments for generalized TL-Align for future works.

%% file: TL-Align.bbl
\begin{thebibliography}{10}\itemsep=-1pt

\bibitem{arnab2021vivit}
Anurag Arnab, Mostafa Dehghani, Georg Heigold, Chen Sun, Mario Lu{\v{c}}i{\'c},
  and Cordelia Schmid.
\newblock Vivit: A video vision transformer.
\newblock In {\em ICCV}, pages 6836--6846, 2021.

\bibitem{ba2016layer}
Jimmy~Lei Ba, Jamie~Ryan Kiros, and Geoffrey~E Hinton.
\newblock Layer normalization.
\newblock {\em arXiv}, abs/1607.06450, 2016.

\bibitem{bao2021beit}
Hangbo Bao, Li Dong, and Furu Wei.
\newblock Beit: Bert pre-training of image transformers.
\newblock {\em arXiv preprint arXiv:2106.08254}, 2021.

\bibitem{cai2018cascade}
Zhaowei Cai and Nuno Vasconcelos.
\newblock Cascade r-cnn: Delving into high quality object detection.
\newblock In {\em CVPR}, pages 6154--6162, 2018.

\bibitem{carion2020end}
Nicolas Carion, Francisco Massa, Gabriel Synnaeve, Nicolas Usunier, Alexander
  Kirillov, and Sergey Zagoruyko.
\newblock End-to-end object detection with transformers.
\newblock In {\em ECCV}, pages 213--229, 2020.

\bibitem{chen2021pre}
Hanting Chen, Yunhe Wang, Tianyu Guo, Chang Xu, Yiping Deng, Zhenhua Liu, Siwei
  Ma, Chunjing Xu, Chao Xu, and Wen Gao.
\newblock Pre-trained image processing transformer.
\newblock In {\em CVPR}, pages 12299--12310, 2021.

\bibitem{chen2021transmix}
Jie-Neng Chen, Shuyang Sun, Ju He, Philip Torr, Alan Yuille, and Song Bai.
\newblock Transmix: Attend to mix for vision transformers.
\newblock {\em arXiv preprint arXiv:2111.09833}, 2021.

\bibitem{chen2022principle}
Tianlong Chen, Zhenyu Zhang, Yu Cheng, Ahmed Awadallah, and Zhangyang Wang.
\newblock The principle of diversity: Training stronger vision transformers
  calls for reducing all levels of redundancy.
\newblock In {\em CVPR}, 2022.

\bibitem{cheng2021per}
Bowen Cheng, Alexander~G Schwing, and Alexander Kirillov.
\newblock Per-pixel classification is not all you need for semantic
  segmentation.
\newblock In {\em NeurIPS}, 2021.

\bibitem{chu2021twins}
Xiangxiang Chu, Zhi Tian, Yuqing Wang, Bo Zhang, Haibing Ren, Xiaolin Wei,
  Huaxia Xia, and Chunhua Shen.
\newblock Twins: Revisiting the design of spatial attention in vision
  transformers.
\newblock In {\em NeurIPS}, 2021.

\bibitem{croce2020reliable}
Francesco Croce and Matthias Hein.
\newblock Reliable evaluation of adversarial robustness with an ensemble of
  diverse parameter-free attacks.
\newblock In {\em International conference on machine learning}, pages
  2206--2216. PMLR, 2020.

\bibitem{dai2021dynamic}
Xiyang Dai, Yinpeng Chen, Jianwei Yang, Pengchuan Zhang, Lu Yuan, and Lei
  Zhang.
\newblock Dynamic detr: End-to-end object detection with dynamic attention.
\newblock In {\em ICCV}, pages 2988--2997, 2021.

\bibitem{dai2021up}
Zhigang Dai, Bolun Cai, Yugeng Lin, and Junying Chen.
\newblock Up-detr: Unsupervised pre-training for object detection with
  transformers.
\newblock In {\em CVPR}, pages 1601--1610, 2021.

\bibitem{deng2009imagenet}
Jia Deng, Wei Dong, Richard Socher, Li-Jia Li, Kai Li, and Li Fei-Fei.
\newblock Imagenet: A large-scale hierarchical image database.
\newblock In {\em CVPR}, pages 248--255. Ieee, 2009.

\bibitem{ding2022scaling}
Xiaohan Ding, Xiangyu Zhang, Yizhuang Zhou, Jungong Han, Guiguang Ding, and
  Jian Sun.
\newblock Scaling up your kernels to 31x31: Revisiting large kernel design in
  cnns.
\newblock In {\em CVPR}, 2022.

\bibitem{dosovitskiy2020image}
Alexey Dosovitskiy, Lucas Beyer, Alexander Kolesnikov, Dirk Weissenborn,
  Xiaohua Zhai, Thomas Unterthiner, Mostafa Dehghani, Matthias Minderer, Georg
  Heigold, Sylvain Gelly, et~al.
\newblock An image is worth 16x16 words: Transformers for image recognition at
  scale.
\newblock In {\em ICLR}, 2020.

\bibitem{guo2022visual}
Meng-Hao Guo, Cheng-Ze Lu, Zheng-Ning Liu, Ming-Ming Cheng, and Shi-Min Hu.
\newblock Visual attention network.
\newblock {\em arXiv preprint arXiv:2202.09741}, 2022.

\bibitem{he2022masked}
Kaiming He, Xinlei Chen, Saining Xie, Yanghao Li, Piotr Doll{\'a}r, and Ross
  Girshick.
\newblock Masked autoencoders are scalable vision learners.
\newblock In {\em Proceedings of the IEEE/CVF Conference on Computer Vision and
  Pattern Recognition}, pages 16000--16009, 2022.

\bibitem{hendrycks2021many}
Dan Hendrycks, Steven Basart, Norman Mu, Saurav Kadavath, Frank Wang, Evan
  Dorundo, Rahul Desai, Tyler Zhu, Samyak Parajuli, Mike Guo, et~al.
\newblock The many faces of robustness: A critical analysis of
  out-of-distribution generalization.
\newblock In {\em Proceedings of the IEEE/CVF International Conference on
  Computer Vision}, pages 8340--8349, 2021.

\bibitem{hendrycks2019benchmarking}
Dan Hendrycks and Thomas Dietterich.
\newblock Benchmarking neural network robustness to common corruptions and
  perturbations.
\newblock {\em arXiv preprint arXiv:1903.12261}, 2019.

\bibitem{hendrycks2021natural}
Dan Hendrycks, Kevin Zhao, Steven Basart, Jacob Steinhardt, and Dawn Song.
\newblock Natural adversarial examples.
\newblock In {\em Proceedings of the IEEE/CVF Conference on Computer Vision and
  Pattern Recognition}, pages 15262--15271, 2021.

\bibitem{jiang2021all}
Zi-Hang Jiang, Qibin Hou, Li Yuan, Daquan Zhou, Yujun Shi, Xiaojie Jin, Anran
  Wang, and Jiashi Feng.
\newblock All tokens matter: Token labeling for training better vision
  transformers.
\newblock In {\em NeurIPS}, 2021.

\bibitem{kim2020puzzle}
Jang-Hyun Kim, Wonho Choo, and Hyun~Oh Song.
\newblock Puzzle mix: Exploiting saliency and local statistics for optimal
  mixup.
\newblock In {\em ICML}, pages 5275--5285, 2020.

\bibitem{krause20133d}
Jonathan Krause, Michael Stark, Jia Deng, and Li Fei-Fei.
\newblock 3d object representations for fine-grained categorization.
\newblock In {\em ICCVW}, pages 554--561, 2013.

\bibitem{krizhevsky2009learning}
Alex Krizhevsky, Geoffrey Hinton, et~al.
\newblock Learning multiple layers of features from tiny images.
\newblock 2009.

\bibitem{li2021efficient}
Wenbo Li, Xin Lu, Jiangbo Lu, Xiangyu Zhang, and Jiaya Jia.
\newblock On efficient transformer and image pre-training for low-level vision.
\newblock {\em arXiv preprint arXiv:2112.10175}, 2021.

\bibitem{li2017not}
Xiaoxiao Li, Ziwei Liu, Ping Luo, Chen Change~Loy, and Xiaoou Tang.
\newblock Not all pixels are equal: Difficulty-aware semantic segmentation via
  deep layer cascade.
\newblock In {\em CVPR}, pages 3193--3202, 2017.

\bibitem{lin2014microsoft}
Tsung-Yi Lin, Michael Maire, Serge Belongie, James Hays, Pietro Perona, Deva
  Ramanan, Piotr Doll{\'a}r, and C~Lawrence Zitnick.
\newblock Microsoft coco: Common objects in context.
\newblock In {\em ECCV}, pages 740--755, 2014.

\bibitem{liu2021swin}
Ze Liu, Yutong Lin, Yue Cao, Han Hu, Yixuan Wei, Zheng Zhang, Stephen Lin, and
  Baining Guo.
\newblock Swin transformer: Hierarchical vision transformer using shifted
  windows.
\newblock In {\em ICCV}, 2021.

\bibitem{liu2022convnet}
Zhuang Liu, Hanzi Mao, Chao-Yuan Wu, Christoph Feichtenhofer, Trevor Darrell,
  and Saining Xie.
\newblock A convnet for the 2020s.
\newblock {\em arXiv preprint arXiv:2201.03545}, 2022.

\bibitem{liu2021video}
Ze Liu, Jia Ning, Yue Cao, Yixuan Wei, Zheng Zhang, Stephen Lin, and Han Hu.
\newblock Video swin transformer.
\newblock {\em arXiv preprint arXiv:2106.13230}, 2021.

\bibitem{nilsback2008automated}
Maria-Elena Nilsback and Andrew Zisserman.
\newblock Automated flower classification over a large number of classes.
\newblock In {\em Indian Conference on Computer Vision, Graphics and Image
  Processing}, 2008.

\bibitem{paszke2019pytorch}
Adam Paszke, Sam Gross, Francisco Massa, Adam Lerer, James Bradbury, Gregory
  Chanan, Trevor Killeen, Zeming Lin, Natalia Gimelshein, Luca Antiga, et~al.
\newblock Pytorch: An imperative style, high-performance deep learning library.
\newblock In {\em NIPS}, pages 8026--8037, 2019.

\bibitem{recht2019imagenet}
Benjamin Recht, Rebecca Roelofs, Ludwig Schmidt, and Vaishaal Shankar.
\newblock Do imagenet classifiers generalize to imagenet?
\newblock In {\em International Conference on Machine Learning}, pages
  5389--5400. PMLR, 2019.

\bibitem{russakovsky2015imagenet}
Olga Russakovsky, Jia Deng, Hao Su, Jonathan Krause, Sanjeev Satheesh, Sean Ma,
  Zhiheng Huang, Andrej Karpathy, Aditya Khosla, Michael Bernstein, et~al.
\newblock Imagenet large scale visual recognition challenge.
\newblock {\em IJCV}, 115(3):211--252, 2015.

\bibitem{strudel2021segmenter}
Robin Strudel, Ricardo Garcia, Ivan Laptev, and Cordelia Schmid.
\newblock Segmenter: Transformer for semantic segmentation.
\newblock In {\em ICCV}, 2021.

\bibitem{tang2021image}
Yehui Tang, Kai Han, Jianyuan Guo, Chang Xu, Yanxi Li, Chao Xu, and Yunhe Wang.
\newblock An image patch is a wave: Quantum inspired vision mlp.
\newblock In {\em CVPR}, 2022.

\bibitem{tolstikhin2021mlp}
Ilya~O Tolstikhin, Neil Houlsby, Alexander Kolesnikov, Lucas Beyer, Xiaohua
  Zhai, Thomas Unterthiner, Jessica Yung, Andreas Steiner, Daniel Keysers,
  Jakob Uszkoreit, et~al.
\newblock Mlp-mixer: An all-mlp architecture for vision.
\newblock {\em NeurIPS}, 34, 2021.

\bibitem{touvron2021resmlp}
Hugo Touvron, Piotr Bojanowski, Mathilde Caron, Matthieu Cord, Alaaeldin
  El-Nouby, Edouard Grave, Gautier Izacard, Armand Joulin, Gabriel Synnaeve,
  Jakob Verbeek, et~al.
\newblock Resmlp: Feedforward networks for image classification with
  data-efficient training.
\newblock {\em arXiv preprint arXiv:2105.03404}, 2021.

\bibitem{touvron2021training}
Hugo Touvron, Matthieu Cord, Matthijs Douze, Francisco Massa, Alexandre
  Sablayrolles, and Herv{\'e} J{\'e}gou.
\newblock Training data-efficient image transformers \& distillation through
  attention.
\newblock In {\em ICML}, pages 10347--10357, 2021.

\bibitem{touvron2022deit}
Hugo Touvron, Matthieu Cord, and Herv{\'e} J{\'e}gou.
\newblock Deit iii: Revenge of the vit.
\newblock {\em arXiv preprint arXiv:2204.07118}, 2022.

\bibitem{uddin2020saliencymix}
AFM Uddin, Mst Monira, Wheemyung Shin, TaeChoong Chung, Sung-Ho Bae, et~al.
\newblock Saliencymix: A saliency guided data augmentation strategy for better
  regularization.
\newblock {\em arXiv preprint arXiv:2006.01791}, 2020.

\bibitem{verma2019manifold}
Vikas Verma, Alex Lamb, Christopher Beckham, Amir Najafi, Ioannis Mitliagkas,
  David Lopez-Paz, and Yoshua Bengio.
\newblock Manifold mixup: Better representations by interpolating hidden
  states.
\newblock In {\em ICML}, pages 6438--6447, 2019.

\bibitem{walawalkar2020attentive}
Devesh Walawalkar, Zhiqiang Shen, Zechun Liu, and Marios Savvides.
\newblock Attentive cutmix: An enhanced data augmentation approach for deep
  learning based image classification.
\newblock {\em arXiv preprint arXiv:2003.13048}, 2020.

\bibitem{wang2021pyramid}
Wenhai Wang, Enze Xie, Xiang Li, Deng-Ping Fan, Kaitao Song, Ding Liang, Tong
  Lu, Ping Luo, and Ling Shao.
\newblock Pyramid vision transformer: A versatile backbone for dense prediction
  without convolutions.
\newblock In {\em ICCV}, 2021.

\bibitem{wei2022activemlp}
Guoqiang Wei, Zhizheng Zhang, Cuiling Lan, Yan Lu, and Zhibo Chen.
\newblock Activemlp: An mlp-like architecture with active token mixer.
\newblock {\em arXiv preprint arXiv:2203.06108}, 2022.

\bibitem{rw2019timm}
Ross Wightman.
\newblock Pytorch image models.
\newblock \url{https://github.com/rwightman/pytorch-image-models}, 2019.

\bibitem{yang2022recursivemix}
Lingfeng Yang, Xiang Li, Borui Zhao, Renjie Song, and Jian Yang.
\newblock Recursivemix: Mixed learning with history.
\newblock {\em arXiv preprint arXiv:2203.06844}, 2022.

\bibitem{yun2019cutmix}
Sangdoo Yun, Dongyoon Han, Seong~Joon Oh, Sanghyuk Chun, Junsuk Choe, and
  Youngjoon Yoo.
\newblock Cutmix: Regularization strategy to train strong classifiers with
  localizable features.
\newblock In {\em ICCV}, pages 6023--6032, 2019.

\bibitem{zhang2018mixup}
Hongyi Zhang, Moustapha Cisse, Yann~N Dauphin, and David Lopez-Paz.
\newblock mixup: Beyond empirical risk minimization.
\newblock In {\em ICLR}, 2018.

\bibitem{zheng2021rethinking}
Sixiao Zheng, Jiachen Lu, Hengshuang Zhao, Xiatian Zhu, Zekun Luo, Yabiao Wang,
  Yanwei Fu, Jianfeng Feng, Tao Xiang, Philip~HS Torr, et~al.
\newblock Rethinking semantic segmentation from a sequence-to-sequence
  perspective with transformers.
\newblock In {\em CVPR}, pages 6881--6890, 2021.

\bibitem{zhou2019semantic}
Bolei Zhou, Hang Zhao, Xavier Puig, Tete Xiao, Sanja Fidler, Adela Barriuso,
  and Antonio Torralba.
\newblock Semantic understanding of scenes through the ade20k dataset.
\newblock {\em IJCV}, 127:302--321, 2019.

\bibitem{zhu2020deformable}
Xizhou Zhu, Weijie Su, Lewei Lu, Bin Li, Xiaogang Wang, and Jifeng Dai.
\newblock Deformable detr: Deformable transformers for end-to-end object
  detection.
\newblock In {\em ICLR}, 2020.

\end{thebibliography}
